\documentclass[3p,twocolumn]{elsarticle}

\usepackage{framed,multirow}

\usepackage{amssymb}
\usepackage{latexsym}

\usepackage{url}
\usepackage{xcolor}
\usepackage{adjustbox}
\usepackage{amsmath}
\definecolor{newcolor}{rgb}{.8,.349,.1}
\usepackage{appendix}

\makeatletter
\def\ps@pprintTitle{%
	\let\@oddhead\@empty
	\let\@evenhead\@empty
	\def\@oddfoot{\footnotesize\itshape 
 Published in
Computer Vision and Image Understanding. DOI: https://doi.org/10.1016/j.cviu.2023.103763
\hfill \today}%
	\let\@evenfoot\@oddfoot}
\makeatother

\journal{Computer Vision and Image Understanding}

\begin{document}

\thispagestyle{empty}

\begin{frontmatter}

\title{Streaming Egocentric Action Anticipation: \\an Evaluation Scheme and Approach}

\author[1,2]{Antonino Furnari\corref{mycorrespondingauthor}}
\cortext[mycorrespondingauthor]{Corresponding author}
\ead{antonino.furnari@unict.it}

\author[1,2]{Giovanni Maria Farinella\corref{}}
\ead{giovanni.farinella@unict.it}
\address[1]{Department of Mathematics and Computer Science, University of Catania, Italy}
\address[2]{Next Vision s.r.l., Italy}

\begin{abstract}
Egocentric action anticipation aims to predict the future actions the camera wearer will perform from the observation of the past.
While predictions about the future should be available before the predicted events take place, most approaches do not pay attention to the computational time required to make such predictions.
As a result, current evaluation schemes assume that predictions are available right after the input video is observed, i.e., presuming a negligible runtime, which may lead to overly optimistic evaluations.
We propose a streaming egocentric action evaluation scheme which assumes that predictions are performed online and made available only after the model has processed the current input segment, which depends on its runtime.
To evaluate all models considering the same prediction horizon, we hence propose that slower models should base their predictions on temporal segments sampled ahead of time.
Based on the observation that model runtime can affect performance in the considered streaming evaluation scenario, we further propose a lightweight action anticipation model based on feed-forward 3D CNNs which is optimized using knowledge distillation techniques with a novel past-to-future distillation loss.
Experiments on the three popular datasets EPIC-KITCHENS-55, EPIC-KITCHENS-100 and EGTEA Gaze+ show that i) the proposed evaluation scheme induces a different ranking on state-of-the-art methods as compared to classic evaluations, ii) lightweight approaches tend to outmatch more computationally expensive ones,
and iii) the proposed model based on feed-forward 3D CNNs and knowledge distillation outperforms current art in the streaming egocentric action anticipation scenario.
\end{abstract}

\begin{keyword}
Action Anticipation, Egocentric Vision, Streaming Perception

\end{keyword}

\end{frontmatter}

\section{Introduction}\label{sec1}
Wearable devices equipped with cameras, such as smart glasses, are recently gaining popularity due to their ability to perceive the world from the user’s point of view and to provide assistance in an effective and natural way~\citep{kanade2012first}.
Among the different tasks investigated in the context of egocentric vision, action anticipation, i.e., predicting a plausible future action which will be performed by the camera wearer from the observation of a past video segment, has attracted a significant amount of attention~\citep{camporese2020knowledge,damen2020epic,dessalene2021forecasting,furnari2020rolling,liu2020forecasting,qi2021self,sener2020temporal,wu2020learning,zhang2020egocentric}.
This attention has been motivated by a general scientific interest in understanding the role of anticipation in human perception and cognition~\citep{ekman2017time,bubic2010prediction} and fostered by the availability of EPIC-KITCHENS~\citep{damen2020epic}, a public benchmark of egocentric videos, with a related action anticipation challenge organized yearly alongside major computer vision conferences\footnote{See \url{https://epic-kitchens.github.io/}.}. 
Moreover, from a practical standpoint, the ability to predict future events is useful when designing technologies aimed to assist humans in their activities where they live and work~\citep{koppula2015anticipating,soran2015generating}.
Indeed, anticipation abilities can enable a wearable device to understand the user’s goals and intentions and provide timely assistance, for example, by selecting appropriate responses to human behavior~\citep{koppula2015anticipating}, by notifying the user about a missing action in a known workflow~\citep{soran2015generating}, or by issuing an alert when the next action likely to be performed is potentially dangerous.

\begin{figure}
    \centering
    \includegraphics[width=\linewidth]{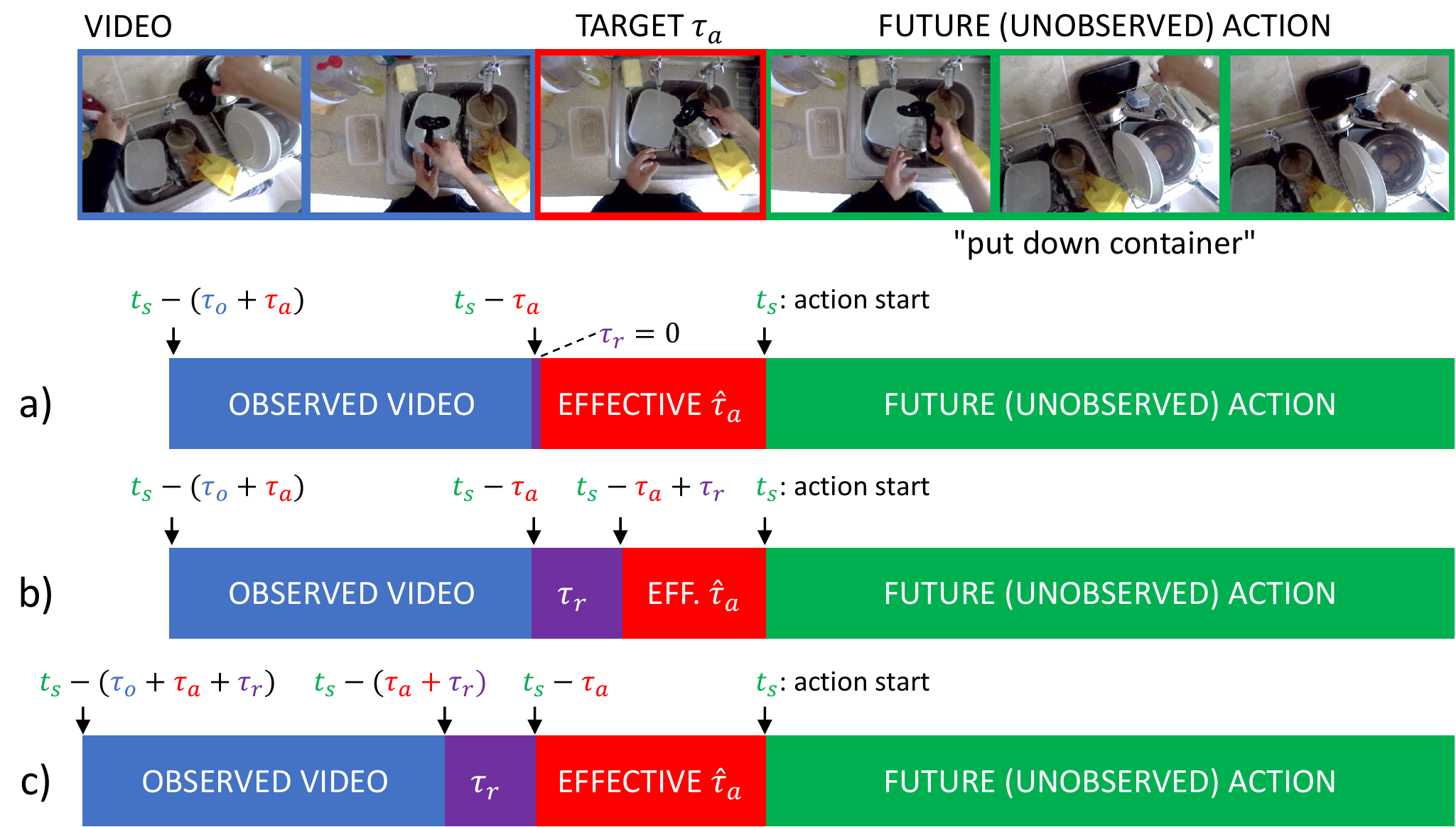}
    \caption{Different schemes to evaluate egocentric action anticipation methods. (a) The ideal commonly used scenario in which the runtime is assumed to be zero. (b) The real-world case in which the runtime is non-negligible. In this case the effective anticipation time is smaller than the target one.
    (c) A fairer evaluation scheme in which the observation is sampled ahead of time to counterbalance the non-negligible runtime and obtain an effective anticipation time equal to the target one. In the figure, $t_s$ denotes the action start timestamp, $\tau_o$ denotes the observation time, $\tau_a$ denotes the anticipation time, $\tau_r$ denotes the runtime, and $\hat \tau_a$ denotes the effective anticipation time.}
    \label{fig:anticipation_runtime}
\end{figure}

While to make action anticipation practically useful the predictions about the future must be available before the action is initiated by the user, previous works have not paid attention to the computational time required to perform such predictions. 
In fact, in general, little attention has been devoted to the influence of runtime on performance in scenarios in which predictions are to be used in real time~\citep{li2020towards}, and egocentric action anticipation approaches have always been evaluated assuming a negligible runtime~\citep{damen2020epic,furnari2020rolling,sener2020temporal,liu2020forecasting,wu2020learning,zhang2020egocentric,qi2021self}, which leads to unfair and overly optimistic evaluations.
Instead, we argue that egocentric action anticipation should be evaluated assuming a “streaming” scenario, in which the anticipation algorithm processes video segments as they become available and outputs predictions after the computational time required by the anticipation model. 
Consider the scheme reported in Figure~\ref{fig:anticipation_runtime}.
An action anticipation model is used to predict the label of an action happening at timestamp $t_s$. 
The model has been trained to anticipate actions happening in $\tau_a$ seconds (referred to as the target anticipation time) by observing a video segment of $\tau_o$ seconds (referred to as observation time).
Current approaches implicitly assume that the model runtime $\tau_r$ is zero and hence that the prediction for a given video segment will be available right after it is passed to the model, neglecting computation time. Hence the model will be evaluated by processing a video observation of temporal bounds $[t_s-(\tau_o+\tau_a), t_s-\tau_a]$ as illustrated in Figure~\ref{fig:anticipation_runtime}(a).
Under the common, but limited, assumption that the model runtime is actually zero, action predictions will be effectively available $\tau_a$ seconds before the beginning of the action, which leads to an effective anticipation time ($\hat \tau_a$) equal to the target one: $\hat \tau_a=\tau_a$.
However, it can be easily seen that if the model runtime is considered to be larger than zero ($\tau_r>0$), predictions about future actions will actually be available at timestamp $t_s-\tau_a+\tau_r$ and hence the effective anticipation time will be smaller than the target one: $\hat \tau_a = \tau_a-\tau_r$ (see Figure~\ref{fig:anticipation_runtime}(b)).
Since different methods are likely characterized by different runtimes, they will also have different effective anticipation times $\hat \tau_a$, which makes comparisons under this classic scheme limited and unfair.

We argue that, in order to evaluate methods more uniformly, the temporal bounds of the observed video should be adjusted at test time to make the effective anticipation time equal to or larger than the target one: $\hat \tau_a \geq \tau_a$.
As highlighted in Figure~\ref{fig:anticipation_runtime}(c), this is possible by shifting the observed video backwards by at least $\tau_r$ seconds, which corresponds to sampling an observed video of temporal bounds $[t_s-(\tau_o+\tau_a+\tau_r), t_s-(\tau_a+\tau_r)]$ for evaluation purposes.
It should be noted that methods with different runtimes will actually observe different video segments, which penalizes methods with a larger runtime, as we expect to happen in a real-world streaming scenario.

Based on the aforementioned observations, in this paper, we propose a new practical evaluation scheme which allows to assess the performance of existing approaches in a streaming fashion by simply quantizing the timestamps at which input video segments are sampled at test time depending on the target anticipation time, the method’s estimated runtime and the length of the video observation.
Since smaller runtimes are more effective when a streaming scenario is assumed, we propose a lightweight egocentric action anticipation model based on simple feed-forward 3D CNNs.
While feed-forward 3D CNNs are conceptually simple and fast, we note that their performance tends to be limited as compared to full-fledged models involving different components.
Taking inspiration from work on knowledge distillation~\citep{hinton2015distilling}, we hence propose to optimize the performance of such models by using a future-to-past knowledge distillation approach which allows to transfer the knowledge from an action recognition model to the target anticipation network.
Experiments on three popular datasets show that 1)~the proposed evaluation scheme induces a different ranking over state-of-the-art methods with respect to non-streaming evaluation schemes, which suggests that current evaluation protocols could be biased and incomplete, 2)~the proposed lightweight method based on 3D CNNs achieves state-of-the-art results in the streaming scenario, which advocates for the viability of knowledge distillation to make learning more data-effective for streaming action anticipation, 3)~lightweight approaches tend to outperform more sophisticated methods requiring different modalities and complex feature extractors in the streaming scenario, which suggests that more attention should be paid to runtime optimization when tackling anticipation tasks.

In sum, the contributions of this work are as follows:
\begin{itemize}
    \item we propose a streaming evaluation scheme which allows to assess egocentric action anticipation methods more fairly and practically. Once the runtime of the approach is estimated, the method can be evaluated by simply quantizing the temporal bounds of the input segment. 
    While this evaluation scheme has been developed for the scenario of egocentric vision in which resources are likely to be limited, it can potentially be applied to action anticipation in a third-person scenario as well;
    \item we perform a benchmark of several state-of-the-art methods and show that the proposed streaming evaluation scheme induces a different ranking with respect to offline evaluation, indicating the limits of current performance assessment protocols;
    \item we propose a lightweight action anticipation model optimized through knowledge distillation and show that it obtains state-of-the-art performance in the streaming scenario, achieving results which are competitive with respect to heavier approaches at a fraction of their computational cost.
\end{itemize}

The rest of this paper is organized as follows: Section~\ref{sec:related} discusses the related work; Section~\ref{sec:streaming_evaluation} presents the proposed streaming evaluation scheme for egocentric action anticipation; Section~\ref{sec:method} introduces the proposed approach based on knowledge distillation;  Section~\ref{sec:experimental_settings} reports the experimental settings; Section~\ref{sec:results} discusses the results; Section~\ref{sec:conclusion} concludes the paper.

\section{Related Work}
\label{sec:related}
Our work is related to previous investigations in the areas of efficient, streaming and online computer vision, egocentric action anticipation, knowledge distillation techniques, and their use to guide future predictions.

\subsection{Efficient, Online and Streaming Computer Vision}
Most image and video understanding tasks, such as image recognition~\citep{he2016deep,simonyan2014very,szegedy2015going,rastegari2016xnor}, object detection~\citep{he2015spatial,lin2017focal,ren2016faster}, action recognition~\citep{carreira2017quo,feichtenhofer2019slowfast,he2015spatial}, and action anticipation~\citep{furnari2020rolling,sener2020temporal,liu2020forecasting}, consider a scenario in which computation is performed offline and hence it is reasonable to assume that computational resources are not limited.
As a result, standard evaluation schemes for these tasks do not generally account for model runtime.
Owing to the practical implications of running computer vision models on hardware in real-world scenarios, past works have proposed computationally-efficient models for image recognition~\citep{tan2019efficientnet,howard2017mobilenets,iandola2016squeezenet}, object detection~\citep{liu2016ssd,redmon2016you,redmon2017yolo9000} and action recognition~\citep{feichtenhofer2020x3d}. 
While these works aimed to jointly optimize for accuracy and computational performance, they have generally been evaluated using standard offline evaluation schemes, while also reporting model runtime or number of FLoating point Operations Per Second (FLOPS).

Another line of research has considered the problem of online video processing, investigating tasks such as online action detection~\citep{de2016online}, early action recognition~\citep{hoai2014max,sadegh2017encouraging}, action anticipation~\citep{damen2020epic,gao2017red,furnari2020rolling,rodin2021predicting} and next active object prediction~\citep{furnari2017next}.
These approaches suggest that a video processing algorithm should be able to make a prediction before a video event is completely observed~\citep{sadegh2017encouraging}, or even before it actually begins~\citep{damen2020epic,furnari2017next}.
Despite considering the usefulness of making predictions in an online fashion, these approaches have not generally paid attention to runtime, requiring algorithms to be able to make a prediction based on partial input, but without posing any constraint on the timeliness of the predictions when computational resources are finite.

Few works have recently considered a ``streaming'' scenario in which algorithms are evaluated also according to the time in which the predictions are available due to the model runtime.
Among these, the VOT 2017 tracking benchmark~\citep{kristan2017visual} included a real-time challenge in which trackers were asked to process video at 20 frames per second, with the most recent prediction being selected if the tracker failed to produce a result before the next frame.
The most systematic investigation in this regard is the one reported in~\citep{li2020towards}, that proposed a meta-benchmark in which object detection algorithms are evaluated assuming a streaming scenario where ground truth annotations available at a given timestamp are compared with the most recent (possibly late) prediction obtained subject to model runtime.

Our investigation takes inspiration from works on efficient computer vision~\citep{feichtenhofer2020x3d,howard2017mobilenets}, online video processing~\citep{de2016online} and streaming perception~\citep{li2020towards}.
Differently from these works, we consider the task of egocentric action anticipation, in which the timeliness of predictions is fundamental to ensure their practical utility. We note that computational efficiency plays a key role in this context due to the mobile nature of wearable devices, which likely leads to constrained computational resources. 
Specifically, we argue that, while current offline evaluation schemes are useful to study algorithmic solutions, they can lead to biased estimations and unfair comparisons when algorithms are supposed to run in real-time on hardware.
We hence propose an easy to implement evaluation scheme which explicitly takes into account model runtime and the streaming nature of video, allowing to compare algorithms in a more homogeneous and practical way.

\subsection{Action Anticipation}
Action anticipation is the task of predicting a future action the camera wearer will likely perform from the observation of a past video segment. The topic has been thoroughly investigated from third-person videos considering goal-oriented scenarios~\citep{pei2011parsing}, human trajectory prediction \citep{kitani2012activity}, prediction of future actions in a dual-actor scenario~\citep{huang2014action}, action anticipation from TV Series video~\citep{lan2014hierarchical}, anticipation for robotic assistance \citep{koppula2015anticipating}, and predicting future representations~\citep{vondrick2016anticipating,gao2017red}. Beyond predicting the next action, some works considered anticipating actions and their starting time~\citep{mahmud2017joint}, as well as predicting longer sequences of future actions~\citep{abu2018will}.

Among more recent works, \cite{zatsarynna2021multi} proposed an efficient architecture based on a stack of a hierarchy of temporal convolutional layers and a multi-modal fusion mechanism to capture pairwise interactions between RGB, flow, and object modalities. 
\cite{zhao2020diverse} tackled the joint anticipation of long-term activity labels together with their lengths using Conditional Adversarial Generative Networks for Discrete Sequences.
\cite{piergiovanni2020adversarial} proposed a method for future anticipation based on adversarial generative grammars which captures temporal dependencies between actions.
\cite{ng2020forecasting} presented an approach to anticipate  actions using a neural machine translation technique based on an encoder-decoder architecture. 
\cite{mehrasa2019variational} introduced a probabilistic generative model for action sequences termed Action Point Process VAE (APP-VAE), a variational auto-encoder able to
capture the distribution of action sequences.
\cite{ke2019time} proposed a time-conditioned method for long-term action anticipation which reduces the accumulation of prediction errors generally due to recurrent models.
\cite{sener2020temporal} designed a model able to predict plausible actions multiple steps into the future in rich natural
language.

\subsection{Egocentric Action Anticipation}
Previous work also considered the problem of anticipating actions from egocentric videos.
Since the introduction of this challenging problem along with the EPIC-KITCHENS dataset~\citep{damen2020epic}, different works have tackled this task~\citep{furnari2020rolling,sener2020temporal,dessalene2021forecasting,liu2020forecasting,girdhar2021anticipative}.
Previous approaches have addressed the task considering baselines designed for action recognition~\citep{damen2020epic}, defining custom losses~\citep{furnari2018leveraging}, modeling the evolution of scene attributes and action over time~\citep{miech2019leveraging}, disentangling the tasks of encoding and anticipation~\citep{furnari2020rolling}, aggregating features over time~\citep{sener2020temporal}, predicting motor attention~\citep{liu2020forecasting}, leveraging contact representations~\citep{dessalene2021forecasting}, mimicking intuitive and analytical thinking~\citep{zhang2020egocentric}, predicting future representations~\citep{wu2020learning}, and modeling attention through transformers~\citep{girdhar2021anticipative}.
While all these approaches have been designed to maximize performance when predicting the future, they have never been evaluated in a streaming scenario. In fact, in general, model runtime has not been reported or even mentioned in past works.

We consider some representatives of these methods and experimentally show that their performance tend to be limited in a streaming scenario, with conceptually simple and lightweight methods outperforming more elaborate and computation-intensive approaches.
We contribute a novel and easy to implement streaming evaluation scheme which can be used to assess performance considering a more realistic and practical scenario, and propose a lightweight model for egocentric action anticipation in resource-constrained settings.

\subsection{Knowledge Distillation}
Knowledge distillation consists in transferring the knowledge from a large and computationally expensive neural network (referred to as the teacher) to a small and generally lightweight network (referred to as the student).
Initial approaches transferred knowledge from the teacher to the student using the logits of the teacher model as a supervisory signal for the student~\citep{hinton2015distilling}.
These methods aimed to minimize the divergence between the probability distributions predicted by the student and the teacher for a given input example, with the goal of providing a richer learning objective as compared to deterministic ground truth labels.
Another line of approaches leveraged the activations of intermediate layers of the teacher to guide the learning of the student~\citep{Huang2017,romero2014fitnets}.
These methods are based on the observation that neural networks tend to learn rich hierarchical representations of their inputs which carry useful information on how an example should be processed.
Other methods further proposed to model the relationships between the outputs of different layers of the network as a way to provide guidance to the student on how to process the input example~\citep{yim2017gift,passalis2020heterogeneous}. Other authors studied the efficacy of knowledge distillation~\citep{cho2019efficacy} with respect to student and teacher architectures. 
Previous works have also explored the use of knowledge distillation techniques to improve action recognition results~\citep{crasto2019mars,stroud2020d3d,liu2019attention}.
We refer the reader to~\citep{gou2021knowledge} for a survey on the topic.

In our work, we use knowledge distillation techniques to optimize the performance of a lightweight action anticipation model and achieve competitive results in the streaming scenario. Differently from the aforementioned works, we do not transfer knowledge from a complex model to a simple one. Rather, we aim to use an action recognition model looking at a future action as the teacher, and a lightweight model looking at a past video snippet as the student. This distillation approach is referred to in this paper as ``future-to-past''. We benchmark different distillation schemes and propose a loss which facilitates knowledge transfer using the teacher to instruct the student on which spatiotemporal features are more discriminative for action anticipation.

\begin{figure*}
    \centering
    \includegraphics[width=0.6\linewidth]{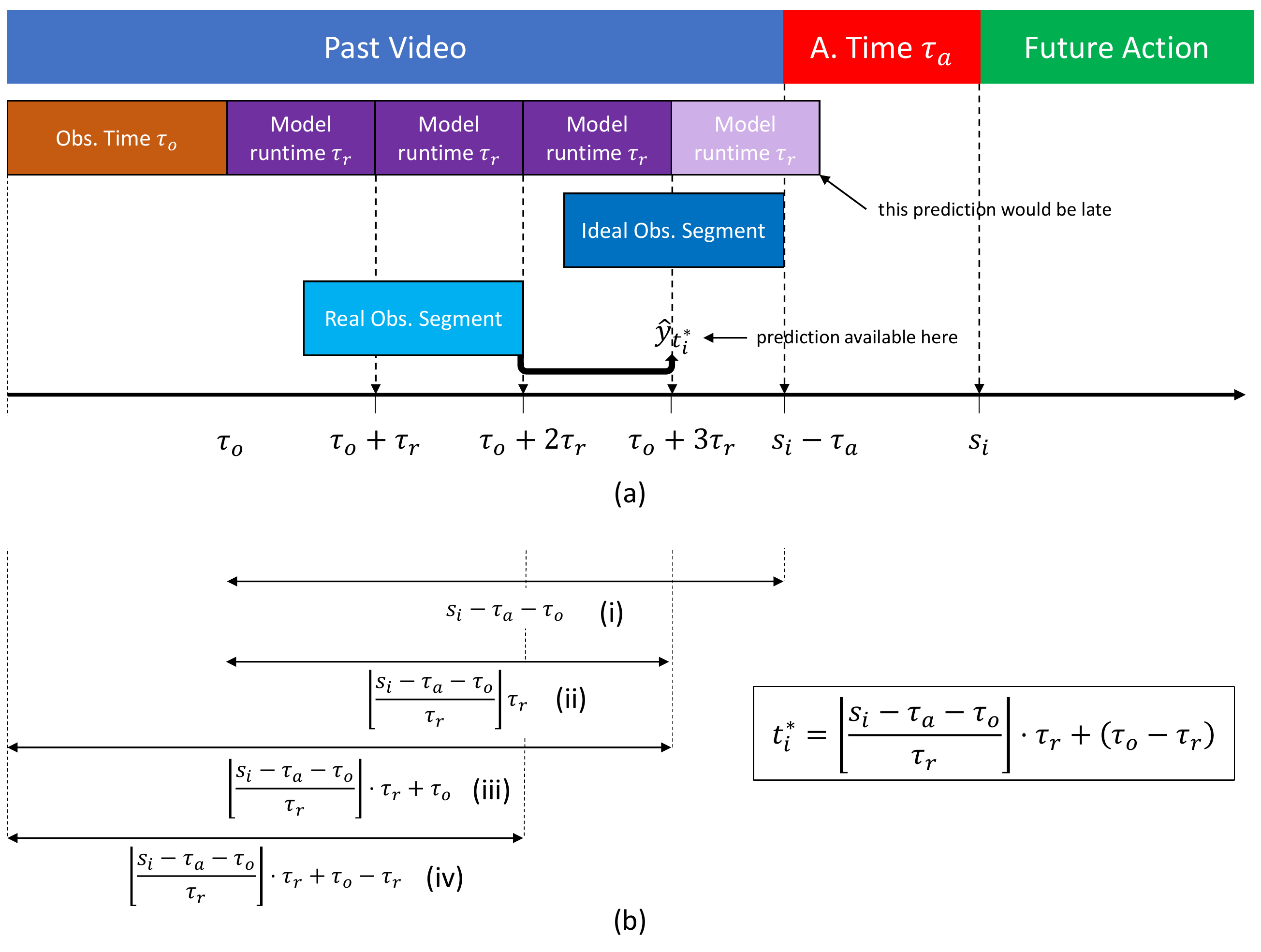}
    \caption{Difference between ideal and real observed segments due to non-negligible model runtimes.
    (a) Due to the observation time and the runtime, a model can process videos only at specific timestamps in a streaming scenario. Specifically, the model has to wait for the observation time $\tau_o$ to form an initial video buffer (left part), then it produces a prediction every $\tau_r$ (runtime) seconds. It should be noted that the model always processes the most recent video segment at the time in which processing is initiated. This induces a difference between the ideal and real observed video segment (right part of the image). (b) Illustration of the quantization formula reported in Eq.~(\ref{eq:quantization}). Once an estimate for the model runtime $\tau_s$ is available, the quantization formula can be used to refine the temporal bounds of the input clips and evaluate methods assuming a streaming scenario with limited computational resources.}
    \label{fig:quantization}
\end{figure*}

\subsection{Knowledge Distillation for Future Prediction}
Few previous works have investigated the use of knowledge distillation to improve action anticipation.
Most works used knowledge distillation implicitly by having the model predict representations of future frames, which were later used to make the actual predictions~\citep{vondrick2016anticipating,gao2017red,wu2020learning}.
Other works have explicitly considered distillation approaches to transfer knowledge from a fixed teacher with label smoothing~\citep{camporese2020knowledge} and from an action recognition model, either pretrained~\citep{tran2019back} or trained jointly with the anticipation model~\citep{Fernando21}.
Other related works have exploited knowledge distillation techniques for the task of early action prediction~\citep{Wang2019}.

Similarly to these previous works, we investigate approaches to transfer the knowledge from an action recognition model to the target anticipation network.
However, differently from previous works, our main objective is to aid the optimization of a lightweight and computationally efficient model.
We set our investigation in the context of egocentric action anticipation with unscripted actions~\citep{damen2020epic}, where predicting future representations is known to be challenging~\citep{furnari2020rolling}.

\section{Proposed Streaming Evaluation Scheme}
\label{sec:streaming_evaluation}
Egocentric action anticipation has been usually evaluated considering a trimmed and offline scenario in which methods are asked to predict a future action by observing a video sampled before the beginning of the action~\citep{damen2020epic,furnari2020rolling,liu2020forecasting}. 
This evaluation protocol is ``trimmed'' because the model is evaluated only at timestamps for which labels are available, it is ``offline'' because the model runtime is assumed not to affect the validity of the predictions.
We consider a ``streaming'' scenario in which the model continuously processes input video segments and makes predictions about the future, but still a ``trimmed'' one in which models are evaluated only one second before the beginning of the action. We opted for a trimmed scenario to deviate as little as possible from the standard definition of short-term egocentric action anticipation~\citep{damen2020epic}, which allows us to better investigate the impact of assuming a streaming scenario. The reader is referred to \citep{rodin2022untrimmed} for a thorough investigation of untrimmed action anticipation.

Specifically, let $V$ be the input video, and let $V_{t_1:t_2}$ denote a video segment starting at timestamp $t_1$ and ending at timestamp $t_2$. Let $\phi$ be the action anticipation algorithm which is designed to process videos of length $\tau_o$ (observation time) and anticipate actions happening after an anticipation time of $\tau_a$ seconds. 
We would like to remark that our protocol does not enforce all methods to adopt the same value for $\tau_o$. Indeed, different methods may opt for different $\tau_o$ values or even adapt this parameter as the predictions are made. In this paper, we kept the values of $\tau_o$ reported by the authors of the compared methods (see the appendix for more details).
At a given timestamp $t$, the algorithm processes the most recent video segment $V_{t-\tau_o:t}$, which is stored in a video buffer containing the most recent frames captured by the camera.
We assume the video buffer to be an ideal component and do not evaluate its latency and computational performance in this paper for simplicity. At each timestamp, the buffer contains the last $\tau_o$ seconds of observed video at the original resolution and framerate. Each algorithm will spatio-temporally sub-sample the video depending on its video processing pipeline.
Let $\tau_r$ be the model runtime, i.e., the time required by the model to process a video segment and output a prediction.
The prediction  $\phi(V_{t-\tau_o:t})$ will hence be available at timestamp $t+\tau_r$, for which we denote it as $\hat y_{t+\tau_r}=\phi(V_{t-\tau_o:t})$. 
Equivalently, we can denote a prediction available at time $t$ as $\hat y_t =\phi(V_{t-\tau_o-\tau_r:t-\tau_r})$, which makes explicit that, in the presence of a large runtime $\tau_r$, predictions should be made by sampling the input video segment ahead of time in order to preserve a correct anticipation time. This means that, to have a prediction at time $t$ with anticipation time $\tau_a$ and runtime $\tau_r$, the last frame observed by the model will be at timestamp $t-\tau_a -\tau_o-\tau_r$.
We assume that resources are limited and only a single GPU process is allowed at a time. This is a realistic scenario when algorithms are deployed to a mobile device such as a pair of smart glasses. Note that in this case the runtime will likely not be negligible ($\tau_r>0$), and hence the model will not be able to make predictions at every single frame of the video. 
In particular, predictions will be available only at selected timestamps $\tau_o + k \cdot \tau_r,k \in N^+$ (the positive integers). 
This happens because the model first needs to wait for $\tau_o$ seconds to form an initial buffer to process, then it has to wait for the runtime $\tau_r$ to make the next prediction.
As a consequence, there will be a difference between the ideal video segment in offline and trimmed settings and the one actually sampled when processing the video in the streaming modality.
Figure~\ref{fig:quantization}(a) illustrates an example of the temporal quantization of predictions introduced by the proposed streaming action anticipation scenario and the consequent difference between ideal and real observed video segments.

\begin{figure*}[t]
    \centering
    \includegraphics[width=\linewidth]{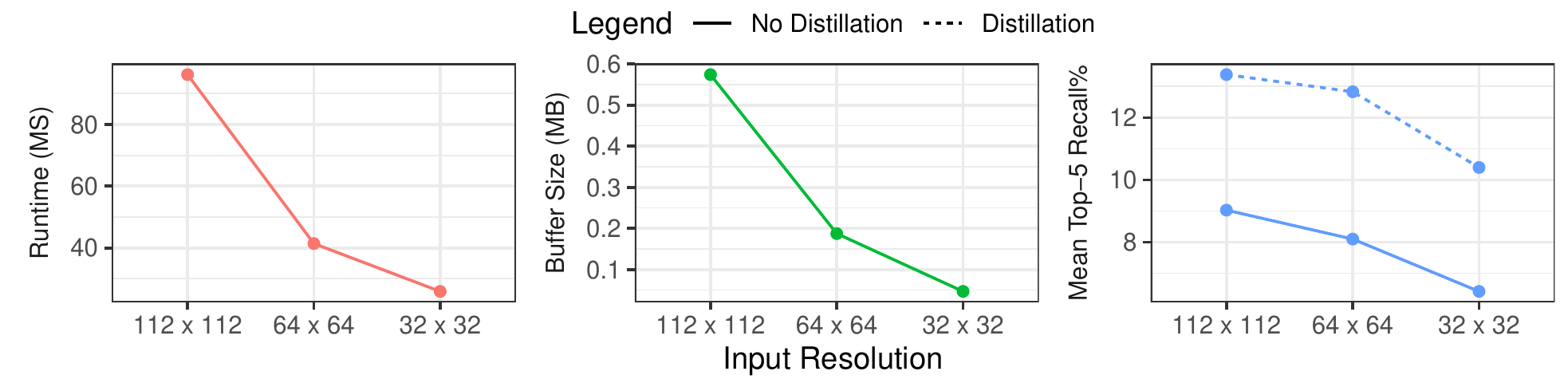}
    \caption{Runtime (left), buffer size (middle) and performance (right) of an R(2+1)D model when trained and tested with videos of different spatial resolutions (all video clips are composed of 16 frames). Lower resolutions allow to reduce model runtime and buffer size, but also affect performance. Knowledge distillation (dashed line in the rightmost plot), allows to mitigate this effect and obtain better performance also in the presence of lower resolutions.}
    \label{fig:runtime_effect}
\end{figure*}

To be consistent with past benchmarks and with the labels available in the current action anticipation datasets, we evaluate the methods only at specific timestamps sampled $\tau_a$ seconds before the beginning of each action.
Specifically, let $A_i=(s_i,e_i,y_i,)$ be the $i^{th}$ labeled action of a test video $V$, where $s_i$ is the action start timestamp, $e_i$ is the action end timestamp, and $y_i$ is the action label. 
Action $A_i$ will be associated to the most recent prediction available $\tau_a$ seconds before the beginning of the action, i.e., at timestamp $s_i-\tau_a$. Note that a prediction might not be available exactly at the required timestamp (see Figure~\ref{fig:quantization}(a)), so we will consider the most recent timestamp at which a prediction is available, that is given by the formula:
\begin{equation}
t_i^*=\left \lfloor \frac{s_i-\tau_a-\tau_o}{\tau_r}\right \rfloor \cdot \tau_r+\tau_o-\tau_r.
\label{eq:quantization}
\end{equation}
Figure~\ref{fig:quantization}(b) illustrates the meaning of the different components of the formula.
$\left \lfloor \frac{s_i-\tau_a-\tau_o}{\tau_r}\right \rfloor$ computes the number of time slots of length $\tau_r$ completely included in the segment of length $s_i-\tau_a-\tau_o$ (denoted as (i) in Figure~\ref{fig:quantization}(b)). Note that subtracting $\tau_o$ is necessary as the first video segment can be processed only after $\tau_o$ seconds. The product $\left \lfloor \frac{s_i-\tau_a-\tau_o}{\tau_r}\right \rfloor \cdot \tau_r$ (denoted as (ii) in Figure~\ref{fig:quantization}(b)) is used to quantize the timestamp. The term $ + \tau_o$ is used to sum the observation time which had been subtracted before (see (iii) in Figure~\ref{fig:quantization}(b)) and the term $-\tau_r$ is used to remove the runtime needed to obtain the prediction (see (iv) in Figure~\ref{fig:quantization}(b)).
The prediction associated to $A_i$ will hence be: $\hat y_{t_i^*}=\phi(V_{t_i^*-\tau_o:t_i^*})$. 
At the beginning of a processed video, $t_i^*$ may be negative. In this case it is not possible to obtain an anticipated prediction for action $A_i$ and hence we predict $\hat y_{t_i^*}$ with a random guess.

Note that, once an estimate for the model runtime $\tau_r$ is available, the presented evaluation method allows to evaluate action anticipation models assuming a more realistic scenario in which computational resources are limited.
Nevertheless, the proposed scheme can be easily implemented following standard evaluation receipies by simply quantizing timestamps at test time using Equation (\ref{eq:quantization}).

\section{Method}
\label{sec:method}
As we noted in Section~\ref{sec:streaming_evaluation}, model runtime influences the performance achieved by action anticipation methods in the streaming scenario. Based on this observation, we propose a lightweight egocentric action anticipation approach based on simple feed-forward 3D CNNs.
Differently from most state-of-the-art action anticipation approaches, which usually involve a multi-modal feature extraction module, followed by a sequence processing component~\citep{furnari2020rolling,sener2020temporal,gao2017red}, feed-forward 3D CNNs are conceptually simple and fast, but, as we will show in the experimental validation, they are harder to optimize for action anticipation and hence they tend to obtain sub-optimal results.
We also note that reducing the resolution of the input video is an effective way to further lower the amount of computational resources needed by the model. 
Figure~\ref{fig:runtime_effect} reports the runtime\footnote{Model runtime is measured using PyTorch and an NVIDIA K80 GPU.}, buffer size (i.e., number of megabytes required to store the input video in the buffer), and anticipation performance of an R(2+1)D model~\citep{tran2018closer} when videos of different resolutions are passed as input.
It is worth noting that smaller resolutions allow to greatly decrease model runtime (Figure~\ref{fig:runtime_effect}-left) and buffer size (Figure~\ref{fig:runtime_effect}-middle), but they also negatively affect model performance (Figure~\ref{fig:runtime_effect}-right).
We argue that this is due to the indirect relationship between input videos and future action labels and to the limited amount of labeled data these models are exposed to. We propose to use knowledge distillation techniques to mitigate this effect (see the dashed line in Figure~\ref{fig:runtime_effect}-right).
In practice, knowledge distillation guides the learning of the feed-forward 3D CNN and also allows to use unlabeled data for training.

In the following sections, we first describe our general training scheme (Section~\ref{sec:training_scheme}), then introduce our proposed distillation loss (Section~\ref{sec:proposed_loss}), and finally give details about the instantiation of the proposed approach (Section~\ref{sec:instantiation}).

\begin{figure}
    \centering
    \includegraphics[width=\linewidth]{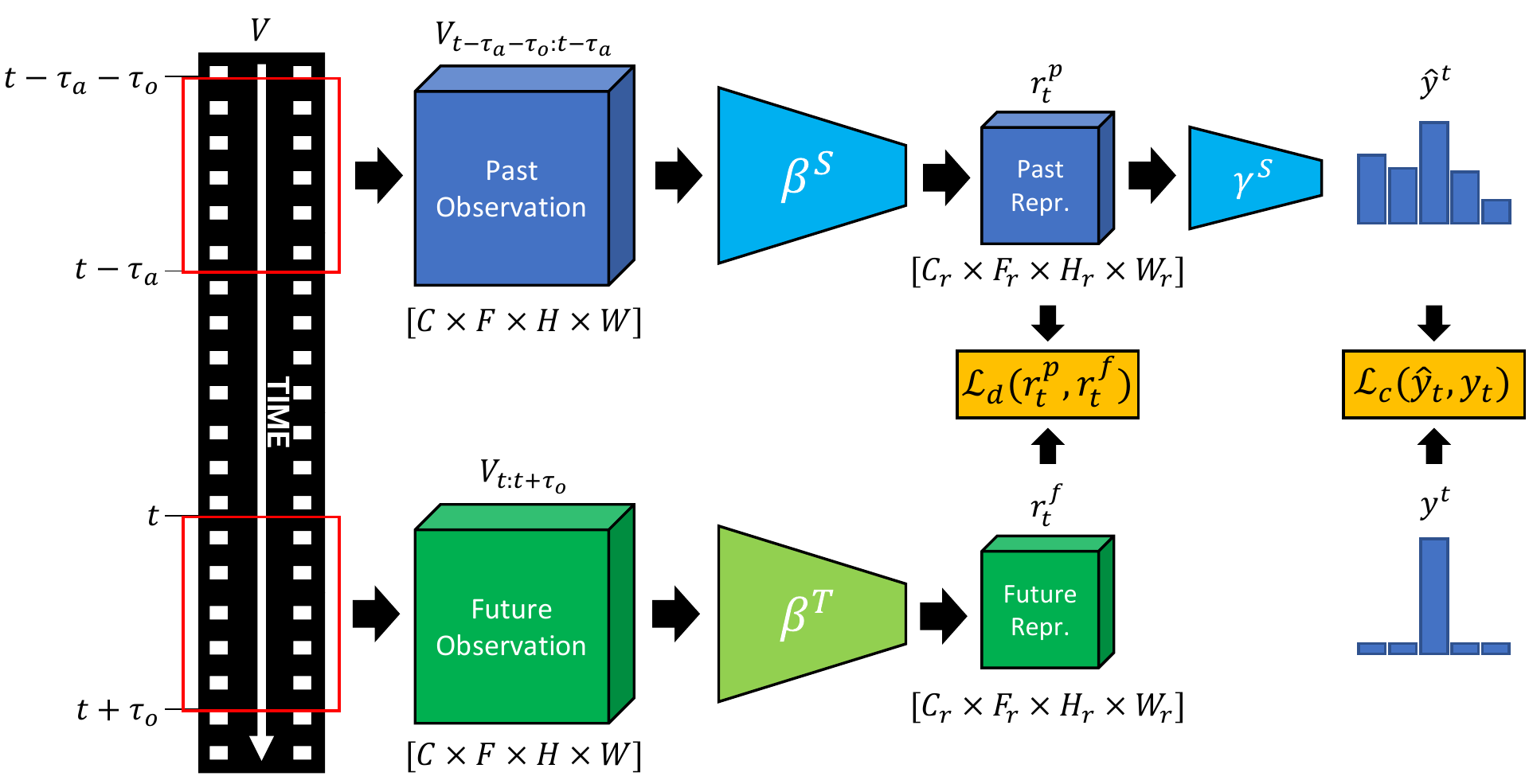}
    \caption{The proposed training procedure. The teacher (bottom) is fed with future video segments $V_{t:t+\tau_o}$, whereas the student (top) processes past video segments $V_{t-\tau_a-\tau_o:t-\tau_a}$. Training is performed both enforcing that the label predicted by the student is correct (classification loss $\mathcal{L}_c$) and that the representations extracted by student and teacher are consistent (distillation loss $\mathcal{L}_d$).}
    \label{fig:training}
\end{figure}

\subsection{Proposed Training Scheme}
\label{sec:training_scheme}
Figure~\ref{fig:training} summarizes the proposed training procedure, which operates over both labeled and unlabeled examples. 
Specifically, given a video $V$ and a timestamp $t$, we form a training example considering a pair of videos including a past observation $V_{t-\tau_a-\tau_o:t-\tau_a}$, a future observation $V_{t:t+\tau_o}$, and a label $y_t$ indicating the class of the action related to the future observation $V_{t:t+\tau_o}$. Note that the future video segment might not be associated to any action (i.e., in the case of unlabeled samples). In this case, the example will be unsupervised, and we will denote its label with $y_t=\varnothing$.
As a teacher, we use a 3D CNN $\phi^T$ trained to perform action recognition from input videos of resolution $C \times F \times H \times W$, where $C$ is the number of channels ($C=3$ for RGB videos), $F$ is the number of frames, $H$ and $W$ are the video frame height and width respectively. The teacher $\phi^T=\beta^T \circ \gamma^T$ is composed of a backbone $\beta^T$ which extracts spatiotemporal representations of resolution $C_r \times F_r \times H_r \times W_r$ (where ``r'' stands for ``representation''), and a classifier $\gamma^T$ which predicts a probability distribution over classes.
The student model $\phi^S=\beta^S \circ \gamma^S$ has the same structure as the teacher $\phi^T$ and it is initialized with the same weights. 
During training, we feed the past observation to the student and obtain both the internal representation of the past segment $r_t^p=\beta^S(V_{t-\tau_a-\tau_o:t-\tau_a })$ and the predicted future action label $\hat y_t = \gamma^S(r_t^p)$. We also extract the representation of the paired future segment $r_t^f=\beta^T(V_{t:t+\tau_o })$.

We hence train the student to both classify the past observation correctly and extract representations which are encouraged to be coherent to the ones of the future segment. To do so, we use the following loss:
\begin{equation}
\label{eq:loss}
    \mathcal{L}=\lambda_d \mathcal{L}_d (r_t^p, r_t^f) + [y_t \neq \varnothing] \lambda_c \mathcal{L}_c (\hat y_t, y_t),
\end{equation}
where $\mathcal{L}_d$ is a distillation loss which enforces that the representations of the past and future segments must be coherent (see Section~\ref{sec:proposed_loss}), $\mathcal{L}_c$ is a classification loss which aims to improve the classification of the past video segment (e.g., cross entropy loss), $\lambda_d$ and $\lambda_c$ are hyperparameters used to regulate the contributions of the two losses, and the term $[y_t\neq \varnothing]$ (Iverson bracket notation) is used to avoid computing the classification loss when the example is unlabeled.

\begin{figure}[t]
    \centering
    \includegraphics[width=\linewidth]{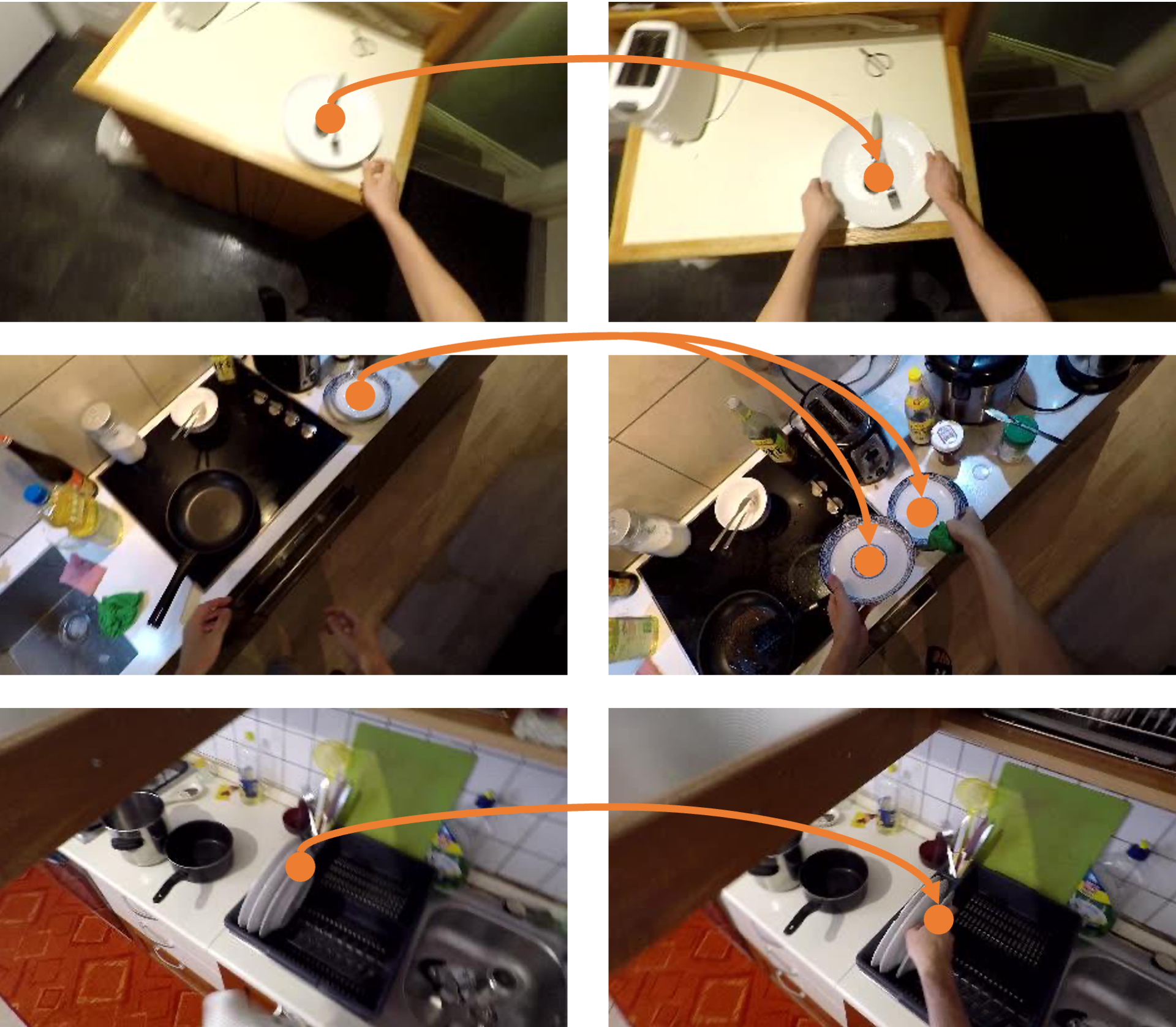}
    \caption{Examples of past (left) and future (right) frames of ``take plate'' actions. The arrows indicate spatiotemporal correspondences of the ``plate'' object across the frames. It is worth noting that, while correspondences are consistently related to ``plate'' objects across different actions, other details of the scene, such as the backgrounds can change. We leverage this observation to define our distillation loss function.}
    \label{fig:take_examples}
\end{figure}

\subsection{Proposed Knowledge Distillation Loss}
\label{sec:proposed_loss}
Differently from classic knowledge distillation approaches, in our training scheme, the teacher and student networks receive different but related inputs (i.e., the past and future video segments), which will likely lead to semantically coherent but spatiotemporally misaligned representations. Consider for instance the case of a ``take plate'' action, as illustrated in Figure~\ref{fig:take_examples}. We would expect ``plate'' features to be available in both the past and future clips, but they will likely be at different spatiotemporal locations.
More specifically, let $r_t^p (i)$ and $r_t^f (i)$ be the $C_r$-dimensional representations obtained at the $i^{th}$ spatiotemporal locations of the tensors $r_t^p$ and $r_t^f$ and let $\mathcal{S}\big(r_t^p (i),r_t^f (i)\big)$ be a measure of similarity between $r_t^p (i)$ and $r_t^f (i)$. Since the two inputs are different, their representations are likely spatiotemporally misaligned. 
Hence a loss trying to maximize directly $\mathcal{S}\big(r_t^p (i),r_t^f (i)\big)  \ \forall i\in{1,\ldots,F_r \cdot H_r \cdot W_r}$ (e.g., the MSE loss) would be sub-optimal.
Ideally, we would like maximize the similarity between corresponding feature pairs $r_t^p(i), r_t^f(\varphi(i))$, where $\varphi$ maps $i$ to the spatiotemporal index corresponding to $i$ in the future representation (e.g., if $r_t^p(i)$ contains ``plate'' features, we expect $r_t^f(\varphi(i))$ to also contain ``plate'' features) . In practice, since the mapping $\varphi$ is unknown, we opt for considering the values of $r_t^p$ and $r_t^f$ as unpaired and maximize all feature pairs: $\mathcal{S}\big(r_t^p (i),r_t^f (j)\big)  \ \forall i,j\in{1,\ldots,F_r \cdot H_r \cdot W_r}$.
While not all feature pairs will be relevant in a single past-future pair, we expect that, in average, relevant pairs will appear consistently in the examples, whereas irrelevant ones will be related to unrelated elements such as the background, as illustrated in Figure~\ref{fig:take_examples}. We hence expect pairs appearing consistently in the examples (such as those related to ``plate'' features) to contribute more to weight updates than irrelevant pairs in our loss formulation.

\begin{figure}
    \centering
    \includegraphics[width=\linewidth]{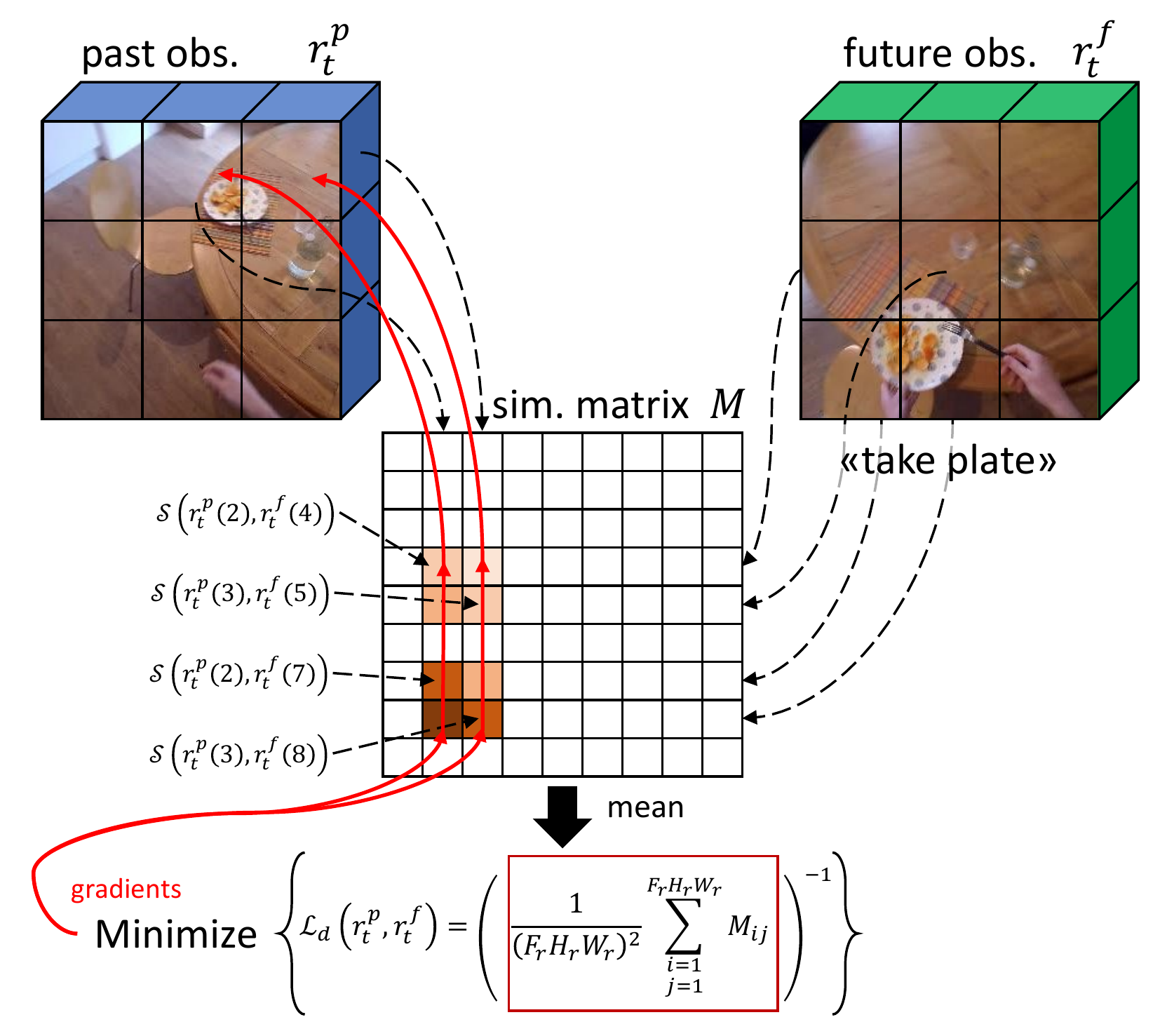}
    \caption{A scheme of the proposed loss in the case of a ``take plate'' future action. The past and future representations are used to compute the similarity matrix between features at all pairs of spatiotemporal locations. Minimizing the loss corresponds to maximizing the values of the similarity matrix, which, in average over many ``take plate'' examples, allows gradients to flow back to the predicted representations, encouraging the student to extract ``plate'' features. In the formula, $r_t^p$ and $r_t^f$ refer to the past and future representations, $F_r$, $H_r$, $W_r$ denote the size of the feature maps, and $M_{ij}$ is the element of the similarity matrix of indices $i$ and $j$.}
    \label{fig:loss}
\end{figure}

We choose $\mathcal{S}$ as the cosine similarity and define a future-to-past similarity matrix $M$ as follows:
\begin{equation}
    M_{ij} = \mathcal{S}\left(r_t^p(i), r_t^f(j)\right) = \frac{r_t^p(i) \cdot r_t^f(j)}{||r_t^p(i)||_2 \cdot ||r_t^f(j)||_2}.
\end{equation}
We hence maximize the values of $M$ by minimizing the following loss:
\begin{equation}
    \mathcal{L}_d(r_t^p,r_t^f) = \left( \frac{1}{(F_r H_r W_r)^2} \sum_{\substack{i=1\\j=1}}^{F_r H_r  W_r} M_{ij} \right)^{-1}.
\end{equation}
The main rationale behind this loss is to encourage the student network to represent semantic concepts in the past observation which are important for action recognition in the future, even if they appear at different spatiotemporal position in the teacher's feature map. Figure~\ref{fig:loss} illustrates the proposed loss.

\subsection{Instantiation of the Proposed Approach}
\label{sec:instantiation}
We instantiate the method described in this paper considering an R(2+1)D backbone~\citep{tran2018closer} for both the student and teacher network.
Specifically, we consider the official PyTorch~\citep{pytorch} implementation of the architecture based on ReseNet18~\citep{he2016deep}.
We choose this model for its good trade-off between performance and computational efficiency.
We consider three different versions of the proposed model which scale their computational efficiency by considering different input sizes: A ``large'' variant DIST-R(2+1)D-L which processes input videos of resolution $3 \times 16 \times 112 \times 112$, a ``medium'' variant DIST-R(2+1)D-M which processes videos of resolution $3 \times 16 \times 64 \times 64$, and a ``small'' variant DIST-R(2+1)D-S which processes videos of resolution $3 \times 32 \times 32$.
Video clips are sampled with a temporal stride of $2$ frames.
We first train the teacher for the action recognition task, then initialize the student with the teacher's weights and train it using the aforementioned training scheme. While training the student, the weights of the teacher are not updated. See the appendix for more details.

\section{Experimental Settings}
\label{sec:experimental_settings}
This  section  discusses  the  datasets  used  for  the  experiments,  the  implementation details of the proposed approach and the compared methods.

\begin{figure*}
    \centering
    \includegraphics[width=\linewidth]{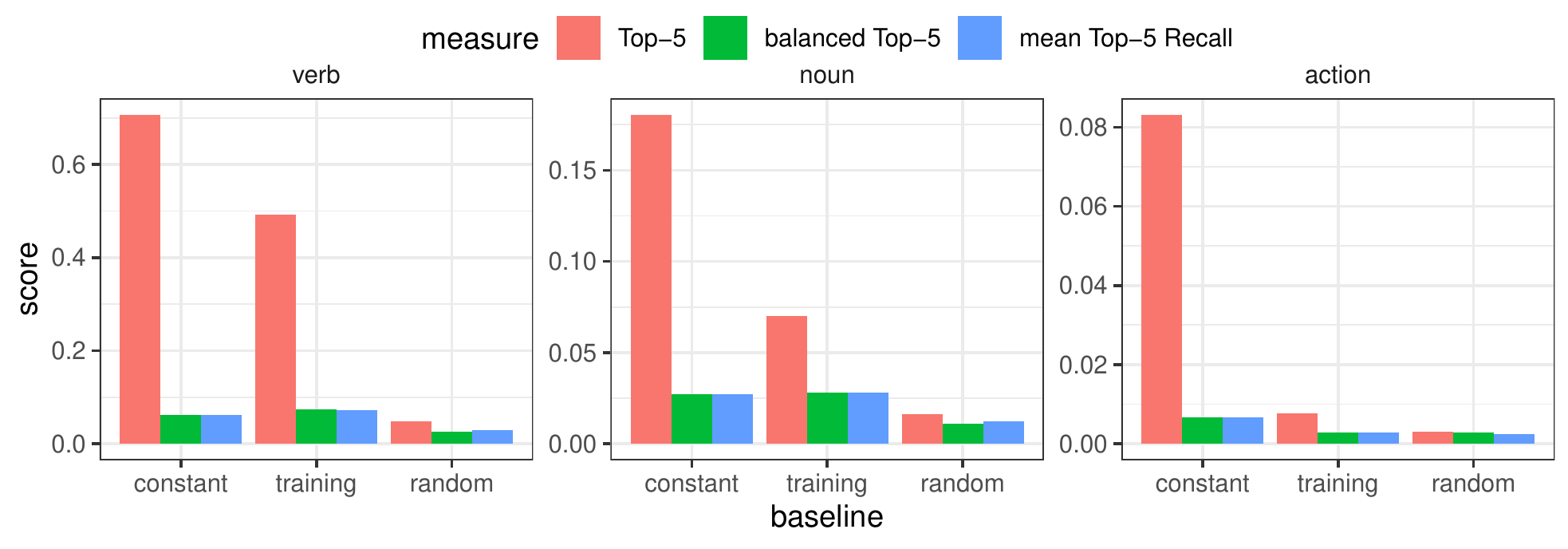}
    \caption{Results obtained by three evaluation measures and three baselines on EPIC-KITCHENS-100. Top-5 Recall tends to overestimate the performance of simple baselines due to class imbalance, whereas mean Top-5 Recall is closer to the results obtained by a theoretical balanced Top-5 accuracy. The diagram highlights that mean Top-5 Recall allows to obtain a more balanced evaluation of algorithm, which leads us to adopt it as the main evaluation measure in this paper.}
    \label{fig:evaluation_measures}
\end{figure*}

\subsection{Datasets}
We perform experiments on three well-known datasets: EPIC-KITCHENS-55~\citep{damen2020epic}, the recently released EPIC-KITCHENS-100~\citep{damen2020rescaling} dataset, and EGTEA Gaze+~\citep{li2021eye}.

EPIC-KITCHENS-55~\citep{damen2020epic} contains $432$ videos collected by $32$ subjects, labeled with $39,595$ action segments including $125$ verbs and $352$ nouns. 
The dataset is divided into a public training set and two private test sets. 
We split the public set as proposed in~\citep{furnari2020rolling} obtaining a training set of $232$ videos and $23,493$ action segments, and a validation set of $40$ videos and $4,979$ segments, on which results are reported.
We consider all unique $(verb, noun)$ pairs appearing in the public set and obtain $2,513$ distinct action classes.

EPIC-KITCHENS-100~\citep{damen2020rescaling} is a recent extension of EPIC-KITCHENS-55 which comprises $700$ videos collected by $37$ different subjects in $45$ kitchens. The videos are labeled with $89,977$ action segments including $97$ verbs and $300$ nouns.
The dataset is split into a public training set, a public validation set and a private test set. We consider all unique $(verb, noun)$ pairs in the training and validation sets and obtain $3806$ unique distinct classes.
Methods are evaluated on EPIC-KITCHENS-100 reporting results on the public validation set.

EGTEA Gaze+~\citep{li2021eye} contains $86$ videos collected by $28$ different subjects, labeled with $10,325$ action segments including $19$ verbs, $51$ nouns and $106$ action classes.
We randomly split the dataset into a training set containing $65$ videos and a test set containing $21$ videos, on which results are reported.
\footnote{Note that we are unable to report results on the three official splits provided by the authors~\citep{li2021eye} as it includes video segments from the same video both in the training and test sets. Indeed, using such split will let the model see unlabeled test videos at training time, thus obtaining overly optimistic performance.}

\subsection{Evaluation Measures}
The proposed streaming egocentric action anticipation evaluation scheme can be implemented considering any performance measure. We base our analysis on the Mean Top-5 Recall (MT5R) measure which has been proposed in~\citep{furnari2018leveraging} and recently adopted in~\citep{damen2020rescaling} as official measure for the egocentric action anticipation task in EPIC-KITCHENS-100.
Similarly to Top-5 accuracy, Mean Top-5 Recall defines a prediction as correct if the ground truth action is included in the top $5$ predictions.
Differently from Top-5 accuracy, Mean Top-5 Recall is a class-aware measure in which performance indicators obtained for each class are averaged to obtain the final score.
As a result, Mean Top-5 Recall is more robust to the class imbalance usually contained in large scale datasets with a long tail distribution~\citep{damen2020rescaling}.

To better understand the limits of standard Top-5 accuracy and the advantages of using Mean Top-5 Recall, in Figure~\ref{fig:evaluation_measures} we compare the performance obtained with three evaluation measures and three different baselines on the verb, noun and action prediction tasks in EPIC-KITCHENS-100.
The considered evaluation measures are the standard Top-5 accuracy, which has been extensively used to evaluate action anticipation algorithms~\citep{damen2020epic}, the mean Top-5 Recall adopted in this work, and an ideal ``balanced'' Top-5 accuracy, which is obtained considering a synthetic test set obtained by sampling $1000$ examples per class.
The considered baselines are 1) a ``constant'' baseline which always predict the verb, noun, and action more frequent in the training set, 2) a ``training'' baseline, which randomly predicts labels following the distributions observed in the training set, and 3) a ``random'' baseline which predicts labels according to a uniform distribution.
Since none of these baselines actually processes the input example, we would expect them to achieve very low performance scores.
Nevertheless, it can be easily seen that both the constant and training baselines achieve very large scores when Top-5 accuracy is used.
This is due to the unbalanced nature of the dataset which makes it easy to optimize Top-5 accuracy by simply considering the distribution of actions in the training set.
Indeed, using a balanced Top-5 evaluation measure would yield more realistic results, with the baseline relying on the training distribution outperforming the constant and random ones, but still achieving overall low scores.
Notably, the adopted mean Top-5 Recall closely follows the scores obtained with the theoretical balanced Top-5 Accuracy. 

While we base our main findings on the Mean Top-5 Recall evaluation measure for the aforementioned motivation, we also report performance according to Top-5 Accuracy for completeness.

\subsection{Compared Methods}
We benchmark different egocentric action anticipation approaches according to both a classic offline evaluation scheme and the proposed streaming protocol described in Section~\ref{sec:streaming_evaluation}.
We choose different representative approaches ranging from well-optimized but computationally expensive methods to the lightweight baselines:
\begin{itemize}
    \item \textbf{Well-optimized approaches}: For this category, we consider the Rolling-Unrolling LSTMs (RULSTM) proposed in~\citep{furnari2020rolling}, and an Encoder-Decoder (ED) architecture based on~\citep{gao2017red}, and implemented as in~\citep{furnari2020rolling}. These approaches make use of LSTMs and consider different input modalities including representations of RGB frames, optical flow, and object-based features. Due to the use of different modalities and to the involved pre-processing, these methods tend to be accurate but computationally expensive.
    \item \textbf{State-of-the-art action recognition baselines}: These methods achieve state-of-the-art results in the action recognition task, but they can be easily adapted to action anticipation. We consider four 3D feed-forward CNNs of varying computational complexity: a computationally expensive I3D~\citep{carreira2017quo} network, the more efficient SlowFast~\citep{feichtenhofer2019slowfast} network, the computationally optimized X3D-XS architecture~\citep{feichtenhofer2020x3d}, and the lightweight R(2+1)D~\citep{tran2018closer} CNN based on Resnet18. As with the proposed method, in order to benchmark the scalability of performance with respect to computational time, we consider three versions of the latter model: R(2+1)D-L, which takes an input of resolution $3 \times 16 \times 112 \times 112$, R(2+1)D-M, which takes an input of resolution $3 \times 16 \times 64 \times 64$, and R(2+1)-S, which takes an input of resolution $3 \times 16 \times 32 \times 32$.
    \item \textbf{Lightweight baselines}: We also consider two approaches designed to be extremely lightweight, albeit likely to be less accurate. Specifically, we consider Temporal Segment Networks (TSN)~\citep{wang2016temporal} and a simple LSTM~\citep{gers1999learning} which takes as input representations of the RGB frames obtained using a BNInception 2D CNN pre-trained for action recognition.
\end{itemize}

\subsection{Implementation Details}
Assuming that resources would be limited on mobile hardware, we estimate all model runtimes using a single NVIDIA K80 GPU.
As done in previous works~\citep{furnari2020rolling}, all models are trained to predict a probability distribution over actions (i.e., $(verb,noun)$ pairs appearing at least once in the dataset). 
We obtain the predicted verb and noun probability distributions by marginalization.
Please see the appendix for additional implementation details.

 \begin{table*}[t]
 \caption{Results on EPIC-KITCHENS-55}
 \label{tab:ek55_results}
 \centering
 \adjustbox{max width=\linewidth}{
 \setlength{\tabcolsep}{2.5pt}
\renewcommand{\arraystretch}{1.2}
 \begin{tabular}{lrr|rrr|rrr|rrr|rrr}
 \multicolumn{3}{}{} &
 \multicolumn{6}{c}{\textbf{Mean Top-5 Recall\%}} &
 \multicolumn{6}{c}{\textbf{Top-5 Accuracy\%}}
 \\
 \multicolumn{3}{}{} & \multicolumn{3}{|c}{\textbf{OFFLINE}} & \multicolumn{3}{|c|}{\textbf{STREAMING}} & \multicolumn{3}{|c}{\textbf{OFFLINE}} & \multicolumn{3}{|c}{\textbf{STREAMING}} \\
   \hline
 \textbf{METHOD} & \textbf{R.TIME} & \textbf{FPS} & \textbf{VERB} & \textbf{NOUN} & \textbf{ACT}. & \textbf{VERB} & \textbf{NOUN} & \textbf{ACT.} & \textbf{VERB} & \textbf{NOUN} & \textbf{ACT.} & \textbf{VERB} & \textbf{NOUN} & \textbf{ACT.}\\ 
   \hline
   RULSTM \citep{furnari2020rolling}  &  724.98 & 1.38 & 21.91 & \underline{27.19} & \bf 14.81 & 19.80 & 24.04 & 11.40 & \bf 79.55 & \bf 51.79 & \bf 35.32 & \bf 76.73 & \underline{46.45} & 28.46 \\
   ED \citep{gao2017red} & 100.56 & 9.94 & 20.97 & 23.35 & 10.60 & 19.87 & 22.41 & 9.60 & 75.83 & 43.03 & 25.59 & 75.08 & 41.84 & 23.93 \\
   \hline
   I3D \citep{carreira2017quo} &  275.26 & 3.63 & 21.50 & 23.68 & 11.77 & 22.08 & 22.63 & 10.83 & 73.95 & 39.82 & 24.05 & 72.92 & 38.99 & 22.60 \\
   SlowFast \citep{feichtenhofer2019slowfast} & 173.73 & 5.76 & 19.97 & 21.10 & 9.87 & 18.53 & 20.96 & 9.31 & 75.99 & 41.29 & 25.22 & 74.78 & 40.62 & 24.05 \\
   X3D-XS \citep{feichtenhofer2020x3d} &  142.50 & 7.02 & 17.49 & 19.47 & 8.54 & 16.56 & 19.25 & 8.02 & 75.88 & 38.64 & 23.25 & 75.26 & 38.39 & 21.72 \\
   R(2+1)D-S \citep{tran2018closer} & 25.90 & 38.61 & 14.31 & 14.66 & 6.42 & 14.69 & 15.11 & 6.20 & 70.23 & 27.94 & 15.53 & 69.94 & 27.88 & 15.33 \\
   R(2+1)D-M \citep{tran2018closer} &  41.41 & 24.15 & 15.31 & 17.09 & 8.10 & 14.52 & 16.82 & 7.89 & 75.48 & 35.62 & 20.29 & 75.14 & 35.17 & 20.06 \\
   R(2+1)D-L \citep{tran2018closer} &  96.14 & 10.40 & 19.02 & 19.35 & 9.03 & 18.59 & 18.55 & 8.49 & 70.63 & 34.00 & 19.24 & 69.90 & 33.10 & 18.69 \\
   \hline
   LSTM \citep{gers1999learning} & 25.96 & 38.52 & 21.59 & 24.62 & 12.35 & 21.22 & 24.80 & 12.18 & \underline{76.85} & 44.88 & 27.68 & \underline{76.57} & 44.80 & 27.98 \\
   TSN \citep{li2020towards} & 19.20 & 52.08 & 19.76 & 21.55 & 10.28 & 19.80 & 21.47 & 10.25 & 74.87 & 40.85 & 25.33 & 74.45 & 40.91 & 25.11 \\
   \hline
   \hline
   DIST-R(2+1)D-S& 25.90 & 38.61 & 20.75 & 22.35 & 10.40 & 20.57 & 22.08 & 10.29 & 72.60 & 36.91 & 21.74 & 72.54 & 36.76 & 21.58 \\
   DIST-R(2+1)D-M &  41.41 & 24.15 & \underline{23.67} & 24.43 & 12.83 & \underline{23.41} & 23.83 & 12.31 & 75.60 & 41.32 & 25.56 & 75.00 & 40.13 & 24.52 \\
   DIST-R(2+1)D-L &  96.14 & 10.40 & 23.49 & 26.30 & 13.38 & 23.33 & 25.43 & 12.70 & 75.93 & 44.68 & 27.78 & 75.85 & 43.43 & 26.45 \\
  \hline
  \hline
  LSTM+DIST-R(2+1)D-S & 51.86 & 19.28 & 22.87 & 26.10 & 13.56 & 22.72 & 25.37 & 12.96 & 74.33 & 45.42 & 29.23 & 74.67 & 44.69 & 28.71 \\
  LSTM+DIST-R(2+1)D-M & 67.37 & 14.84 & 23.51 & 26.11 & \underline{14.35} & 22.65 & \underline{26.60} & \underline{13.55} & 76.67 & 46.93 & 30.94 & 75.56 & 45.78 & \underline{29.60} \\
  LSTM+DIST-R(2+1)D-L & 122.10 & 8.19 & \bf 24.52 & \bf 27.21 & 14.15 & \bf 24.08 & \bf 26.71 & \bf 13.68 & 75.98 & \underline{47.90} & \underline{31.75} & 75.27 & \bf 47.47 & \bf 30.88 \\
  \hline
\end{tabular}}
\end{table*}

\section{Results}
\label{sec:results}
In this section, we report and discuss the results of our experimental analysis. Specifically, Section~\ref{sec:benchmark} presents the benchmark of the different egocentric action anticipation methods on the considered datasets, whereas Section~\ref{sec:ablation} reports ablation results on EPIC-KITCHENS-55 aimed to assess the role of the different design choices of our method and the general effect of knowledge distillation on performance in the considered streaming scenario.

\begin{figure*}
    \centering
    \includegraphics[width=0.85\linewidth]{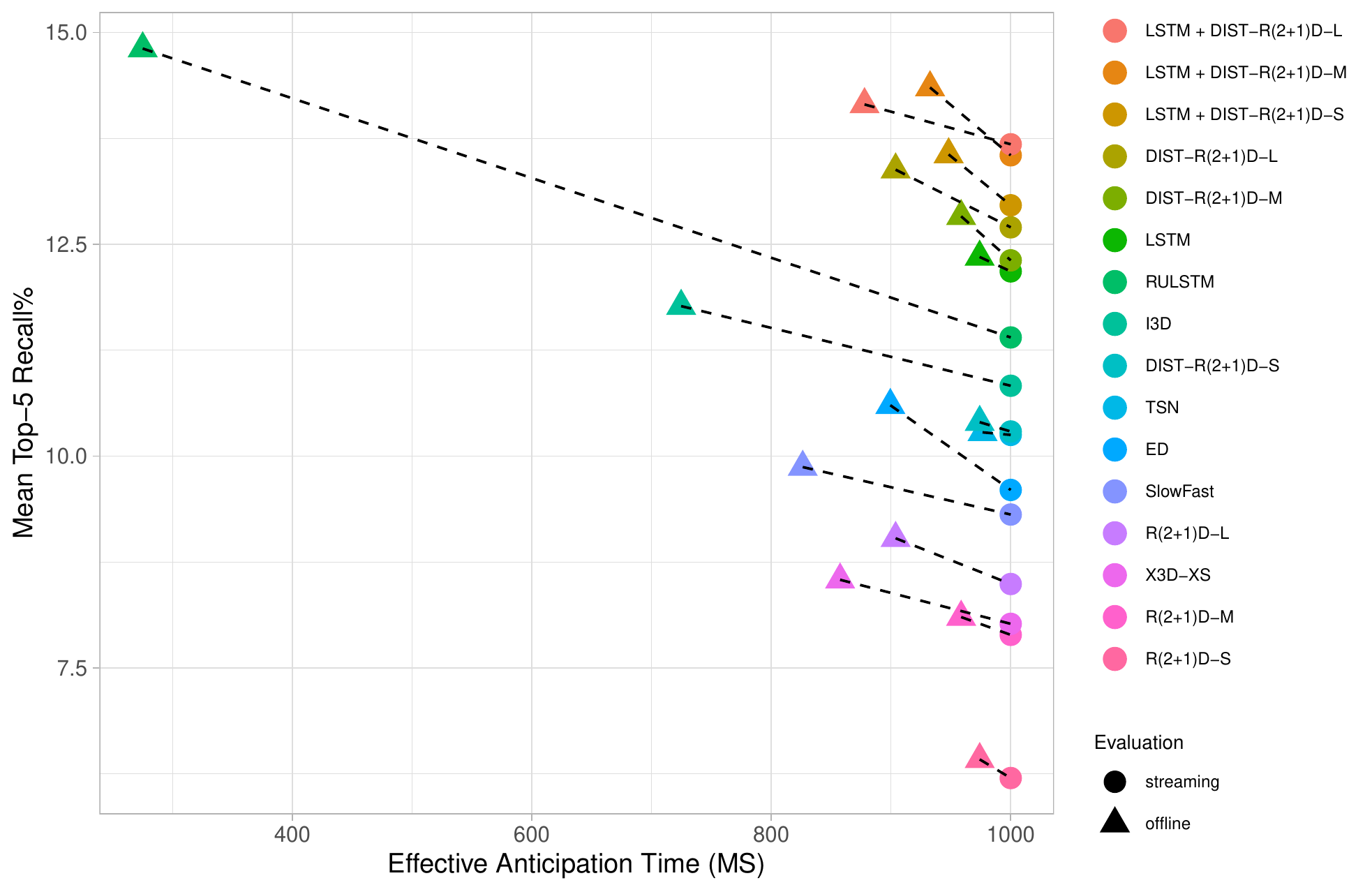}
    \caption{Effect on performance of the streaming egocentric action anticipation evaluation scheme on EPIC-KITCHENS-55. The legend is sorted according to streaming performance. The performance of computationally expensive methods such as RULSTM and I3D tends to be more affected by the streaming scenario, whereas lightweight models such as the proposed DIST-R(2+1)D-L and DIST-R(2+1)D-M are more robust to the streaming evaluation scheme.}
    \label{fig:ek55_adjust}
\end{figure*}

\subsection{Streaming Egocentric Action Anticipation Benchmark}
The streaming egocentric action anticipation benchmark compares methods both considering the standard offline evaluation settings and according to the proposed streaming evaluation scheme. At the same time, the benchmark compares the proposed knowledge distillation approach to prior art.
\label{sec:benchmark}

\subsubsection{EPIC-KITCHENS-55}
Table~\ref{tab:ek55_results} reports the performance of the compared methods according to both the classic offline evaluation scheme (columns $4-6$ and $10-12$) and the proposed streaming evaluation protocol (columns $7-9$, $13-15$).
We report results using Mean Top-5 Recall\%, which is the main evaluation measure considered in this study (columns $4-9$), and Top-5 Accuracy\%, which is reported for completeness (columns $10-15$).
For each measure, we report verb, noun and action (ACT.) performance. Best results per column are reported in bold, whereas second-best results are underlined.
For each method, we also report in columns $2$ and $3$ the runtime in milliseconds (R.TIME), which includes both the time needed for pre-processing (e.g., feature extraction) and the time needed by the model to perform inference, as well as the framerate (FPS), which indicates how frequently the proposed method can be executed given its runtime.

\begin{table*}[t]
\caption{Results on EPIC-KITCHENS-100}
\label{tab:ek100_results}
\centering
\adjustbox{max width=\linewidth}{
\setlength{\tabcolsep}{2.5pt}
\renewcommand{\arraystretch}{1.2}
\begin{tabular}{lrr|rrr|rrr|rrr|rrr}
 \multicolumn{3}{}{} &
 \multicolumn{6}{c}{\textbf{Mean Top-5 Recall\%}} &
 \multicolumn{6}{c}{\textbf{Top-5 Accuracy\%}}
 \\
 \multicolumn{3}{}{} & \multicolumn{3}{|c}{\textbf{OFFLINE}} & \multicolumn{3}{|c|}{\textbf{STREAMING}} & \multicolumn{3}{|c}{\textbf{OFFLINE}} & \multicolumn{3}{|c}{\textbf{STREAMING}} \\
   \hline
 \textbf{METHOD} & \textbf{R.TIME} & \textbf{FPS} & \textbf{VERB} & \textbf{NOUN} & \textbf{ACT}. & \textbf{VERB} & \textbf{NOUN} & \textbf{ACT.} & \textbf{VERB} & \textbf{NOUN} & \textbf{ACT.} & \textbf{VERB} & \textbf{NOUN} & \textbf{ACT.}\\ 
  \hline
  RULSTM \citep{furnari2020rolling} & 724.98 & 1.38 & 27.76 & \bf 30.76 & \underline{14.04} & 23.27 & 27.42 & 11.07 & \bf 79.11 & \bf 55.34 & \bf 36.46 & 74.74 & \underline{49.02} & 29.30 \\
   ED \citep{gao2017red} & 100.56 & 9.94 & 24.43 & 26.59 & 10.60 & 23.23 & 27.41 & 10.49 & 74.28 & 46.95 & 28.96 & 73.57 & 46.47 & 28.35 \\
   \hline
   I3D \citep{carreira2017quo} & 275.26 & 3.63 & 27.58 & 27.60 & 12.79 & 25.44 & 26.07 & 11.37 & 73.30 & 46.17 & 29.29 & 71.46 & 43.76 & 26.71 \\
   SlowFast \citep{feichtenhofer2019slowfast} & 173.73 & 5.76 & 23.86 & 23.03 & 9.90 & 22.91 & 22.45 & 9.60 & 75.68 & 46.29 & 29.53 & 74.70 & 44.83 & 28.17 \\
   X3D-XS \citep{feichtenhofer2020x3d} & 142.50 & 7.02 & 21.47 & 22.49 & 8.69 & 20.61 & 21.95 & 8.02 & 73.73 & 42.29 & 25.93 & 72.84 & 41.48 & 24.71 \\
   R(2+1)D-S \citep{tran2018closer} & 25.90 & 38.61 & 16.88 & 14.92 & 5.19 & 17.02 & 14.23 & 4.93 & 63.53 & 26.81 & 14.82 & 63.02 & 27.04 & 14.66 \\
   R(2+1)D-M \citep{tran2018closer} & 41.41 & 24.15 & 19.49 & 18.93 & 7.57 & 18.95 & 18.71 & 7.03 & 69.59 & 34.94 & 20.37 & 68.72 & 34.50 & 20.08 \\
   R(2+1)D-L \citep{tran2018closer} & 96.14 & 10.40 & 22.11 & 21.91 & 8.59 & 20.80 & 21.27 & 8.09 & 68.96 & 37.41 & 22.87 & 67.69 & 36.89 & 22.09 \\
   \hline
   LSTM \citep{gers1999learning} & 25.96 & 38.52 & 22.94 & 27.35 & 12.49 & 23.34 & 26.99 & 12.38 & 74.72 & 45.69 & 27.43 & 74.80 & 45.29 & 27.30 \\
   TSN \citep{li2020towards} & 19.20 & 52.08 & 21.69 & 21.95 & 8.70 & 21.21 & 21.89 & 8.64 & 74.59 & 44.30 & 27.06 & 74.31 & 43.98 & 26.87 \\
   \hline
   \hline
   DIST-R(2+1)D-S & 25.90 & 38.61 & 20.68 & 21.50 & 7.77 & 20.76 & 21.12 & 7.42 & 71.39 & 37.38 & 22.33 & 71.01 & 37.05 & 21.81 \\
   DIST-R(2+1)D-M & 41.41 & 24.15 & 23.07 & 25.53 & 11.18 & 22.50 & 25.78 & 11.18 & 74.19 & 44.11 & 27.82 & 73.76 & 43.87 & 27.22 \\
   DIST-R(2+1)D-L & 96.14 & 10.40 & 26.40 & 27.54 & 12.87 & 25.58 & 26.71 & 11.83 & 76.09 & 47.97 & 31.29 & 75.01 & 47.06 & 29.81 \\
   \hline
   \hline
   LSTM+DIST-R(2+1)D-S & 51.86 & 19.28 & 27.28 & 27.53 & 11.70 & 26.98 & 26.88 & 11.26 & 74.60 & 45.86 & 30.09 & 73.85 & 45.35 & 29.39 \\
   LSTM+DIST-R(2+1)D-M & 67.37 & 14.84 & \underline{29.34} & 29.40 & 13.53 & \underline{28.25} & \underline{29.35} & \underline{12.79} & 75.90 & 49.42 & 32.67 & \underline{75.23} & 48.87 & \underline{31.71} \\
   LSTM+DIST-R(2+1)D-L & 122.10 & 8.19 & \bf 29.83 & \underline{30.62} & \bf 14.32 & \bf 28.92 & \bf 29.95 & \bf 13.58 & \underline{76.24} & \underline{51.33} & \underline{34.30} & \bf 75.47 & \bf 50.12 & \bf 33.05 \\
  \hline
\end{tabular}}
\end{table*}

As can be noted from Table~\ref{tab:ek55_results}, models with large runtimes are more optimized for prediction performance and hence they tend to outperform competitors in the standard offline evaluation scenario both in the case of Mean Top-5 Recall and Top-5 Accuracy. For instance, ED is outperformed by RULSTM, similarly based on LSTMs but making use of object-based features, which greatly affects runtime ($100.56 ms$ of ED vs $724.98 ms$ of RULSTM). Similarly, I3D outperforms SlowFast and X3D-XS according Mean Top-5 Recall, also thanks to its larger input size ($3 \times 64 \times 224 \times 224$ of I3D vs $3 \times 32 \times 224 \times 224$ of SlowFast and $3 \times 4 \times 160 \times 160$ of X3D-XS), at the cost of a larger runtime ($275.26 ms$ of I3D vs $173.73 ms$ of SlowFast and $142.5 ms$ of X3D XS).
A similar trend is generally followed by R(2+1)D-S, R(2+1)D-M, and R(2+1)D-L, whose performances and runtimes are affected by the input size.
For instance, R(2+1)D-L has the largest runtime among the three methods ($96.14 ms$ of R(2+1)D-L vs $41.41 ms$ of R(2+1)D-M and $25.90 ms$ of R(2+1)D-S) but also the largest offline action Mean Top-5 Recall ($9.03$ vs $8.10$ vs $6.42$).
In general, methods based on simple 3D CNNs tend to perform worse than highly optimized methods such as RULSTM.
Notably, the proposed approach based on knowledge distillation improves the performance of all the three versions of the R(2+1)D networks for both Mean Top-5 Recall and Top-5 Accuracy.
Indeed, adding distillation to the models based on R(2+1)D allows to boost Action Mean Top-5 Recall in the offline scenario by $+3.63$ ($10.40$ of DIST-R(2+1)D-S vs $6.42$ of R(2+1)D-S), $+4.73$ ($12.83$ of DIST-R(2+1)D-M vs $8.10$ of R(2+1)D-M), and $+4.35$ ($13.38$ of DIST-R(2+1)D-L vs $9.03$ of R(2+1)D-L). Similarly, Top-5 Accuracy is increased by $+6.21$ ($21.74$ of DIST-R(2+1)D-S vs $15.53$ of R(2+1)D-S), $+5.27$ ($25.56$ of DIST-R(2+1)D-M vs $20.29$ of R(2+1)D-M), and $+8.54$ ($27.78$ of DIST-R(2+1)D-L vs $19.24$ of R(2+1)D-L).

It is worth noting that runtime greatly affects performance when the streaming evaluation scenario is considered.
Indeed, while the performances of all methods decrease when passing from offline to streaming evaluation (compare column $6$ with column $9$ and column $10$ with column $13$), methods characterized by smaller runtimes experience a smaller performance gap.
For instance, RULSTM has an offline Action Mean Top-5 Recall of $14.81$, which drops down to $11.40$ in the streaming scenario ($-3.41$), whereas TSN, which is the fastest approach, has an offline Action Mean Top-5 Recall of $10.28$ which drops down to $10.25$ in the streaming scenario (only $-0.03$).
Interestingly, the performance of the lightweight LSTM approach (runtime of $25.96ms$) is barely affected when passing from an offline to an online evaluation scheme ($12.35$ offline action performance vs $12.18$ streaming action performance). Similar considerations hold in the case of Top-5 Accuracy.

The bottom part of Table~\ref{tab:ek55_results} reports the performance of methods obtained by combining the LSTM with the proposed models based on knowledge distillation.
Results point out that the obtained LSTM+DIST-R(2+1)D-S, LSTM+DIST-R(2+1)D-M, and LSTM+DIST-R(2+1)D-L achieve good results at small computational costs.
For example, LSTM+DIST-R(2+1)D-L achieves a streaming Action Mean Top-5 Recall of $13.68$ with a runtime of $122.10ms$ and a still usable framerate of $8.19fps$, whereas LSTM+DIST-R(2+1)D-M achieves a similar streaming Action Mean Top-5 Recall of $13.55$, while running much faster with a runtime of $67.37ms$ and a framerate of $14.84fps$. Similar observations can be made in the case of Top-5 Accuracy. In this case, LSTM+DIST-R(2+1)D+L achieves the best Noun and Action Top-5 Accuracy, with a Verb Top-5 Accuracy performance comparable to the one obtained by the much slower RULSTM ($1.38fps$).

Notably, the offline and streaming evaluation schemes tend to induce different rankings of the methods, which suggests that classic offline evaluation offers only a limited picture when algorithms are to be deployed to real hardware with limited resources.
This is better illustrated in Figure~\ref{fig:ek55_adjust}, which represents methods with respect to their effective anticipation time (x axis) and performance measured in action Mean Top-5 Recall\% (y axis). 
For each method, the figure reports both the offline performance (triangles) and the streaming performance (circles).
Dashed lines connect 2D points related to the same method evaluated in the streaming and offline scenarios.
As previously observed, methods with larger runtimes (triangles on the left part of the figure) tend to experience large performance gaps, which induce a different ranking among the methods.
It i sworth noting that, the best performing approaches in the streaming scenario are the proposed models based on large and medium R(2+1)D networks, also in combination with the LSTM model.

\begin{table*}[t]
\caption{Results on EGTEA Gaze+}
\label{tab:egtea_results}
\centering
\adjustbox{max width=\linewidth}{
\setlength{\tabcolsep}{2.5pt}
\renewcommand{\arraystretch}{1.2}
\begin{tabular}{lrr|rrr|rrr|rrr|rrr}
 \multicolumn{3}{}{} &
 \multicolumn{6}{c}{\textbf{Mean Top-5 Recall\%}} &
 \multicolumn{6}{c}{\textbf{Top-5 Accuracy\%}}
 \\
 \multicolumn{3}{}{} & \multicolumn{3}{|c}{\textbf{OFFLINE}} & \multicolumn{3}{|c|}{\textbf{STREAMING}} & \multicolumn{3}{|c}{\textbf{OFFLINE}} & \multicolumn{3}{|c}{\textbf{STREAMING}} \\
   \hline
 \textbf{METHOD} & \textbf{R.TIME} & \textbf{FPS} & \textbf{VERB} & \textbf{NOUN} & \textbf{ACT}. & \textbf{VERB} & \textbf{NOUN} & \textbf{ACT.} & \textbf{VERB} & \textbf{NOUN} & \textbf{ACT.} & \textbf{VERB} & \textbf{NOUN} & \textbf{ACT.}\\ 
  \hline
  RULSTM \citep{furnari2020rolling} & 724.98 & 1.38 & 78.53 & \underline{71.11} & \underline{59.79} & 71.17 & 60.12 & 49.17 & 91.77 & 77.64 & \underline{67.68} & 86.90 & 68.01 & 56.13 \\
   ED \citep{gao2017red} & 100.56 & 9.94 & 73.54 & 64.99 & 52.78 & 73.79 & 62.68 & 52.23 & 90.17 & 73.11 & 61.02 & 89.90 & 72.04 & 59.68 \\
   \hline
   I3D \citep{carreira2017quo} & 275.26 & 3.63 & 77.37 & 65.59 & 52.53 & 73.15 & 60.16 & 48.31 & 88.59 & 68.90 & 56.64 & 85.17 & 64.08 & 51.08 \\
   SlowFast \citep{feichtenhofer2019slowfast} & 173.73 & 5.76 & 68.08 & 58.56 & 44.16 & 66.45 & 55.74 & 42.45 & 90.13 & 72.31 & 59.07 & 88.54 & 69.78 & 57.02 \\
   X3D-XS \citep{feichtenhofer2020x3d} & 142.50 & 7.02 & 70.79 & 61.73 & 45.16 & 68.42 & 58.81 & 44.93 & 89.43 & 70.02 & 56.45 & 88.26 & 67.96 & 54.58 \\
   R(2+1)D-S \citep{tran2018closer} & 25.90 & 38.61 & 63.18 & 46.85 & 34.52 & 63.02 & 47.26 & 34.78 & 84.85 & 57.62 & 45.65 & 84.14 & 57.39 & 44.86 \\
   R(2+1)D-M \citep{tran2018closer} & 41.41 & 24.15 & 68.38 & 56.83 & 42.52 & 67.52 & 54.61 & 41.48 & 87.28 & 65.90 & 53.04 & 86.95 & 64.22 & 51.40 \\
   R(2+1)D-L \citep{tran2018closer} & 96.14 & 10.40 & 74.19 & 62.82 & 49.04 & 70.12 & 58.95 & 47.06 & 89.20 & 70.11 & 57.58 & 87.14 & 66.70 & 54.49 \\
   \hline
   LSTM \citep{gers1999learning} & 25.96 & 38.52 & 77.80 & 67.68 & 58.15 & 77.16 & 67.02 & 56.52 & 91.81 & 75.91 & 66.23 & \underline{91.30} & 75.54 & 64.73 \\
   TSN \citep{li2020towards} & 19.20 & 52.08 & 69.86 & 58.89 & 46.49 & 70.01 & 58.58 & 45.99 & 89.52 & 71.23 & 59.07 & 89.38 & 70.53 & 58.75 \\
   \hline
   \hline
   DIST-R(2+1)D-S & 25.90 & 38.61 & 67.16 & 53.54 & 41.13 & 66.61 & 54.19 & 40.66 & 88.82 & 66.88 & 54.72 & 88.54 & 66.84 & 54.07 \\
   DIST-R(2+1)D-M & 41.41 & 24.15 & 76.29 & 64.84 & 49.82 & 73.80 & 65.51 & 49.87 & 91.35 & 74.28 & 60.90 & 91.11 & 73.85 & 59.96 \\
   DIST-R(2+1)D-L & 96.14 & 10.40 & 78.08 & 67.60 & 55.72 & 74.19 & 66.01 & 54.30 & \bf 92.66 & 76.29 & 65.39 & \bf 91.44 & 73.95 & 63.24 \\
   \hline
   \hline
   LSTM+DIST-R(2+1)D-S & 51.86 & 19.28 & 75.45 & 66.21 & 55.45 & 75.73 & 66.34 & 54.17 & 90.83 & 76.71 & 66.00 & 90.88 & 76.10 & 65.20 \\
   LSTM+DIST-R(2+1)D-M & 67.37 & 14.84 & \underline{80.75} & 70.45 & 57.95 & \bf 79.48 & \underline{69.79} & \underline{57.33} & 91.81 & \underline{78.86} & 67.26 & 90.97 & \underline{77.60} & \underline{66.56} \\
   LSTM+DIST-R(2+1)D-L & 122.10 & 8.19 & \bf 80.77 & \bf 71.30 & \bf 60.17 & \underline{77.17} & \bf 69.99 & \bf 59.08 & \underline{92.28} & \bf 79.37 & \bf 69.13 & 91.16 & \bf 77.78 & \bf 67.73 \\
  \hline
\end{tabular}}
\end{table*}

\subsubsection{EPIC-KITCHENS-100}
Table~\ref{tab:ek100_results} reports the results on EPIC-KITCHENS-100.
Similarly to what observed in the case of EPIC-KITCHENS-55 (Table~\ref{tab:ek55_results}), methods characterized by larger runtimes tend to outperform lightweight models in the offline scenario. For instance, RULSTM obtains an Action Mean Top-5 Recall of $14.04$, versus $12.49$ obtained by a simple LSTM.
Also in this case, adding knowledge distillation allows to improve the results of the R(2+1)D models.
In particular DIST-R(2+1)D-S obtains an offline action accuracy of $7.77$ (+$2.58$ with respect to R(2+1)D-S), DIST-R(2+1)D-M obtains an offline action accuracy of $11.49$ (+$3.92$ with respect to R(2+1)D-M), and DIST-R(2+1)D-L obtains an offline action accuracy of $12.87$ (+$4.28$ with respect to R(2+1)D-L). Similar improvements are observed in the case of Action Top-5 Accuracy: $+7.51$ ($22.33$ of DIST-R(2+1)D-S vs $14.82$ of R(2+1)D-S), $+7.45$ ($27.82$ of DIST-R(2+1)D-M vs $20.37$ of R(2+1)D-M), and $+8.42$ ($31.29$ of DIST-R(2+1)D-L vs $22.87$ of R(2+1)D-L).
Similarly to Table~\ref{tab:ek55_results}, methods with a smaller runtime experience a smaller performance gap when passing from an offline to a streaming evaluation scenario. For instance, LSTM passes from an Action Mean Top-5 Recall of $12.49$ in offline settings to $12.38$ in the streaming scenario (only $-0.11$).
The best results are achieved combining the LSTM to the three proposed approaches based on knowledge distillation. This allows to obtain methods with a reasonable runtime and framerate which outperform state-of-the-art approaches.
For instance, the proposed LSTM+DIST+R(2+1)D-L achieves a streaming Action Mean Top-4 Recall of $13.58$, which is a $+2.51$ improvement with respect to RULSTM ($11.07$ streaming Action Mean Top-5 Recall) at a runtime of $122.10ms$ (vs $724.98ms$ of RULSTM) and a framerate of $8.19fps$ (vs $1.38fps$ of RULSTM). Similar considerations apply also when comparing methods with Top-4 Accuracy: the proposed LSTM+DIST-R(2+1)D-L approach is second-best according to the offline evaluation and best according to the online evaluation.
Our approach is scalable. Indeed, LSTM+DIST-R(2+1)D-M is much faster (framerate of $14.84fps$) at the cost of a small drop in performance ($12.79$ vs $13.58$ streaming Action Mean Top-5 Recall).

\subsubsection{EGTEA Gaze+}
Table~\ref{tab:egtea_results} reports the results on the EGTEA Gaze+ dataset.
Coherently with previous findings, more optimized methods such as RULSTM achieve better offline performance at the cost of a larger runtime, whereas when passing to the streaming scenario, lightweight methods have an advantage. For instance, LSTM achieves an offline Action Mean Top-5 Recall of $58.15$, which is lower ($-1.64$) as compared to the offline Action Mean Top-5 Recall of RULSTM ($59.79$).
However, in the streaming scenario, the LSTM scores an Action Mean Top-5 Recall of $56.52$, which is higher ($+7.35$) than $49.17$, the action performance of RULSTM.
The proposed distillation technique improves the performance of R(2+1)D-based methods also on this dataset.
In particular, DIST-R(2+1)D-S improves the offline Action Mean Top-5 Recall of R(2+1)D-S by $+6.61$, DIST-R(2+1)D-M improves the offline action performance of R(2+1)D-M by $+7.3$, and DIST-R(2+1)D-L improves the offline action performance of R(2+1)D-L by $+6.68$. Similar grains are observed also in the case of Top-5 Accuracy.
The best performance in the streaming scenario is obtained also in this case by LSTM+DIST-R(2+1)D-L, which obtains a streaming Action Mean Top-5 Recall of $59.08$ with a runtime of $122.10ms$ and a framerate of $8.19fps$.
A faster alternative is given by LSTM+DIST-R(2+1)D-M, which achieves a streaming Action Mean Top-5 Recall of $57.33$ at a runtime of $67.37ms$ and a framerate of $14.84fps$. LSTM+DIST-R(2+1)-L is the second-best method for offline Action Top-5 Accuracy and the best method for streaming Action Top-5 Accuracy.

\subsubsection{General Findings} Results on all datasets highlight that the proposed streaming evaluation scenario offers a complementary perspective with respect to classic offline evaluation.
Indeed, method ranking changes depending on model runtime when the streaming scenario is considered.
This suggests that attention should be paid to the evaluation protocol, especially when models have to be deployed in a real scenario.
In particular, all experiments have shown that lightweight approaches such as a simple combination of a 2D CNN and an LSTM can outperform more complex approaches in the streaming scenario, thanks to their reduced runtime.
Experiments also shown that the proposed knowledge distillation techniques can be effectively used to improve the performance of simple 3D feed-forward networks, allowing to greatly improve performance while keeping a restrained runtime.
Additionally, due to their small runtime, the proposed models can be easily fused with other lightweight methods such as LSTMs to further improve performance. 

\subsection{Ablation Study}
\label{sec:ablation}
In this section, we report experiments aimed at highlighting some important properties of the investigated knowledge distillation approach in relation to the deployment of action anticipation models in a streaming scenario. In particular, Section~\ref{sec:data_efficient} shows that, in our context, knowledge distillation acts as a regularizer allowing to train the model on a mix of labeled and unlabeled examples. Section~\ref{sec:ablation_scheme} shows that the proposed ``future-to-past'' distillation scheme in which a recognition model is distilled into an anticipation one is advantageous with respect to classic ``big-to-small'' distillation schemes in which larger models are distilled into smaller ones. Section~\ref{sec:ablation_loss} compares the proposed loss with respect to other distillation losses, highlighting that using cost functions which do not assume spatio-temporally aligned past and future representations allows to improve results.

The experiments reported in this section have been performed on EPIC-KITCHENS-55 considering the R(2+1)D-M model, which offers a good trade-off between performance and computational efficiency.
Indeed, Table~\ref{tab:ek55_results} shows that while R(2+1)D-L obtains best single-model absolute streaming performance, R(2+1)D-M is $43\%$ faster than R(2+1)D-L (R(2+1)D-M runs at $24.15 fps$ vs $10.40 fps$ of R(2+1)D-L) while achieving very comparable performance in the streaming scenario ($12.70\%$ Mean Top-5 Recall of R(2+1)D-L vs $12.31\%$ Mean Top-5 Recall of R(2+1)D-M).
All results are reported considering Mean Top-5 Recall and according to the proposed streaming evaluation scenario.

\subsubsection{Knowledge Distillation as Data-Efficient Training}
\label{sec:data_efficient}
We argue that the proposed future-to-past knowledge distillation approach acts as a regularizer which allows the model to learn how to anticipate future actions more effectively.
The regularizer has two effects: 1) It allows to guide the learning of the anticipation model using the action recognition teacher to highlight discriminative features, 2) It allows to leverage unlabeled examples for training, which effectively increases the amount of training data.
To verify these two claims, in Table~\ref{tab:ablation_data} we compare the performance of the R(2+1)D-M model when trained with different amounts of training data and in the presence or not of knowledge distillation.
Specifically, we consider the following sets of data: 1) Supervised - training examples are obtained sampling video clips exactly one second before the beginning of the action, without any temporal augmentations; 2) Augmented - training examples are obtained with temporal augmentation, i.e., we sample videos randomly and assign them the label of a future video segment located after one second. If no label appears in the future segment or if the label of the observed and future segments match, the example is discarded. Note that this set also includes all supervised examples. 3) All - we consider all possible video clips, both labeled and unlabeled. Note that this set can only be used when knowledge distillation techniques are considered.

\begin{table}[t]
\caption{Effect of Data and Knowledge Distillation (Mean Top-5 Recall\%)}
\label{tab:ablation_data}
\centering
\adjustbox{max width=\linewidth}{
\setlength{\tabcolsep}{2.5pt}
\renewcommand{\arraystretch}{1.2}
\begin{tabular}{lllrrr}
  \hline
\textbf{Data} & \textbf{Examples} & \textbf{Dist.} & \textbf{VERB} & \textbf{NOUN} & \textbf{ACT.} \\ 
  \hline
  Supervised & $23,493$ & No & 14.52 & 16.82 & 7.89 \\ 
  Supervised & $23,493$ & Yes & 19.48 & 21.3 & 9.98 \\
  \hline
  Augmented & $993,899$ & No & 18.25 & 21.51 & 9.20 \\ 
  Augmented & $993,899$ & Yes & \underline{20.80} &	\underline{23.82} &	\underline{10.37} \\ 
  \hline
  All & $3,584,241$ & Yes & \textbf{23.41} &	\textbf{23.83} &	\textbf{12.31} \\ 
  \hline
\end{tabular}}
\end{table}

The results in Table~\ref{tab:ablation_data} show that just adding knowledge distillation techniques, keeping the amount of training data fixed, always allows to improve performance, which suggests that the guiding effect of knowledge distillation helps improving generalization. For instance, training the model with knowledge distillation using only supervised data allows to obtain a $+2.09$ in action performance as compared to training without knowledge distillation ($9.98$ vs $7.89$). 
Just augmenting the number of training data has a similar regularizing effect. Indeed, training the model with the augmented data allows to obtain an action score of $10.37$, which is a $+1.48$ with respect to the $7.89$ obtained considering only supervised data.
Using both approaches allows to boost performance by $+4.42$ with respect to the base model ($12.31$ using all data and knowledge distillation vs $7.89$ using only supervised data and no knowledge distillation).
These results highlight how knowledge distillation has the effect of making training more data-efficient allowing to increase the number of training examples and better exploit the available data.

\subsubsection{Knowledge Distillation Scheme}
\label{sec:ablation_scheme}
In Table~\ref{tab:ablation_scheme}, we compare the proposed future-to-past knowledge distillation scheme to the much more common ``big-to-small'' knowledge distillation scheme in which knowledge is transferred from a computationally expensive model to a lightweight one~\citep{hinton2015distilling}.
Specifically, we use an R(2+1)-L model trained for action anticipation as a teacher and an R(2+1)-M model as a student.
Differently from the proposed training scheme, in this case, both teacher and student get the past observation as the input video segment.
The rest of the training process is the same as ``future-to-past''.
Since our loss has not been optimized for this big-to-small scheme, we consider three common knowledge distillation losses.
The first one, is the original Knowledge Distillation (KD) loss proposed in~\citep{hinton2015distilling}. 
The second one computes the Mean Squared Error (MSE) loss assuming that the extracted feature maps are spatio-temporally aligned.
The third one, first performs Global Average Pooling (GAP) of both (teacher and student) feature maps to obtain spatio-temporally agnostic representations, then applies the MSE loss.
The results show that, while all approaches allow to improve over the baseline with no distillation (first row), the proposed future-to-past scheme is much more effective than the classic big-to-small one.

\begin{table}[t]
\caption{Comparison of Distillation Schemes (Mean Top-5 Recall\%)}
\label{tab:ablation_scheme}
\centering
\adjustbox{max width=\linewidth}{
\setlength{\tabcolsep}{2.5pt}
\renewcommand{\arraystretch}{1.2}
\begin{tabular}{llrrr}
  \hline
\textbf{Dist. Scheme} & \textbf{Dist. Loss} & \textbf{VERB} & \textbf{NOUN} & \textbf{ACT.} \\ 
  \hline
  No Distillation & - & 14.52 & 16.82 & 7.89 \\ 
  \hline
  big-to-small & KD~\citep{hinton2015distilling} & 20.62 & 22.85 & 10.65\\ 
  big-to-small & GAP + MSE & 20.29 & 22.31 & 10.80 \\ 
  big-to-small & MSE & \underline{20.93} & \underline{24.42} & \underline{11.09} \\ 
 \hline
  future-to-past & Proposed & \textbf{23.41} & \textbf{23.83} & \textbf{12.31} \\ 
   \hline
 \end{tabular}}
 \end{table}

\subsubsection{Knowledge Distillation Loss}
\label{sec:ablation_loss}
We finally compare the performance of the proposed loss with respect to other known and baseline knowledge distillation losses in Table~\ref{tab:ablation_loss}. Specifically, we compare with respect to the original Knowledge Distillation (KD) loss~\citep{hinton2015distilling}, the Variational Information Distillation (VID) loss~\citep{ahn2019variational}, the Maximum Mean Discrepancy (MMD) loss~\citep{Huang2017}, and the ``Back To The Future'' (BTTF) loss proposed in~\citep{tran2019back}. We also consider the MSE and GAP + MSE baselines as defined in Section~\ref{sec:ablation_scheme}.
As can be seen from Table~\ref{tab:ablation_loss} shows that all distillation approaches allow to improve results over the baseline model which does not consider knowledge distillation (first row). 
Among the compared approaches, the proposed one achieves best action performance ($12.31$), whereas verb and noun performances are second-best, but comparable to GAP+MSE and MSE.
Interestingly, combining GAP with MSE, allows to improve average performance ($19.78$ vs $19.13$), with a particular advantage over verbs ($23.53$ vs $22.3$).
In average, the proposed approach outperforms all competitors.

\begin{table}[t]
\caption{Comparison of Distillation Loss Functions (Mean Top-5 Recall\%)}
\label{tab:ablation_loss}
\centering
\adjustbox{max width=0.85\linewidth}{
\setlength{\tabcolsep}{2.5pt}
\renewcommand{\arraystretch}{1.2}
\begin{tabular}{lrrrr}
  \hline
\textbf{Method} & \textbf{VERB} & \textbf{NOUN} & \textbf{ACT.} & \textbf{AVG.} \\ 
  \hline
  No Distillation & 14.52 &	16.82 &	7.89 &	13.08 \\ 
  \hline
  KD \citep{hinton2015distilling} & 20.69	& 23.47 &	10.89 &	18.35\\ 
  VID \citep{ahn2019variational} & 21.3 &	24.25 &	11.34 &	18.96 \\ 
  MMD \citep{Huang2017} & 20.87 &	23.49 &	11.28 &	18.55 \\ 
  BTTF \citep{tran2019back} & 20.95 &	\underline{23.65} &	11.61 &	18.74 \\ 
  \hline
  MSE & 22.3 &	23.22 &	11.88 &	19.13\\ 
  GAP + MSE & \textbf{23.53} &	\underline{23.65} &	\underline{12.17} &	\underline{19.78} \\ 
  \hline
  Proposed & \underline{23.41} & \textbf{23.83} & \textbf{12.31} & \textbf{19.85} \\ 
  \hline
\end{tabular}}
\end{table}

\subsection{Qualitative Examples}
\begin{figure*}
    \centering
    \includegraphics[width=\linewidth]{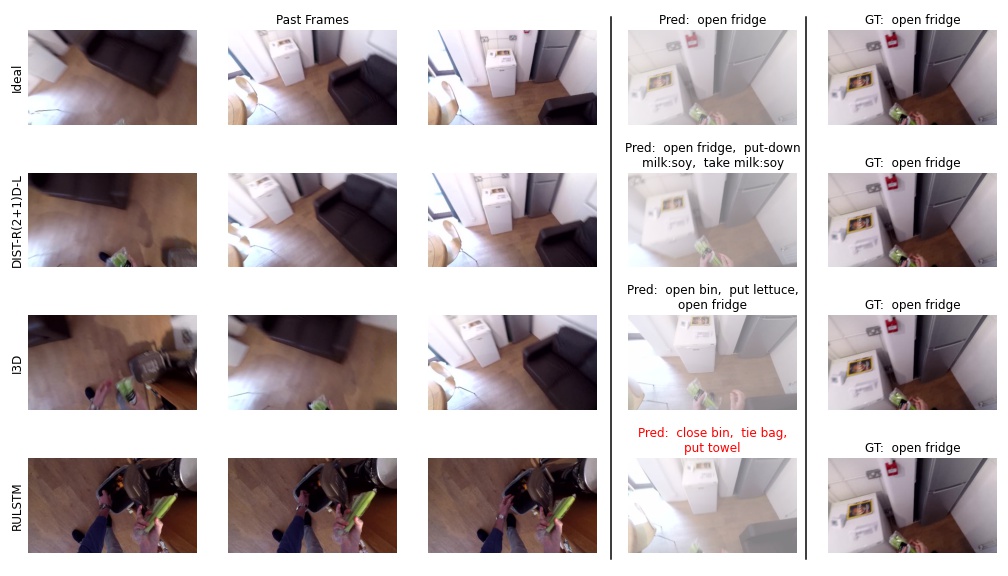}
    
    \vspace{5mm}
    \includegraphics[width=\linewidth]{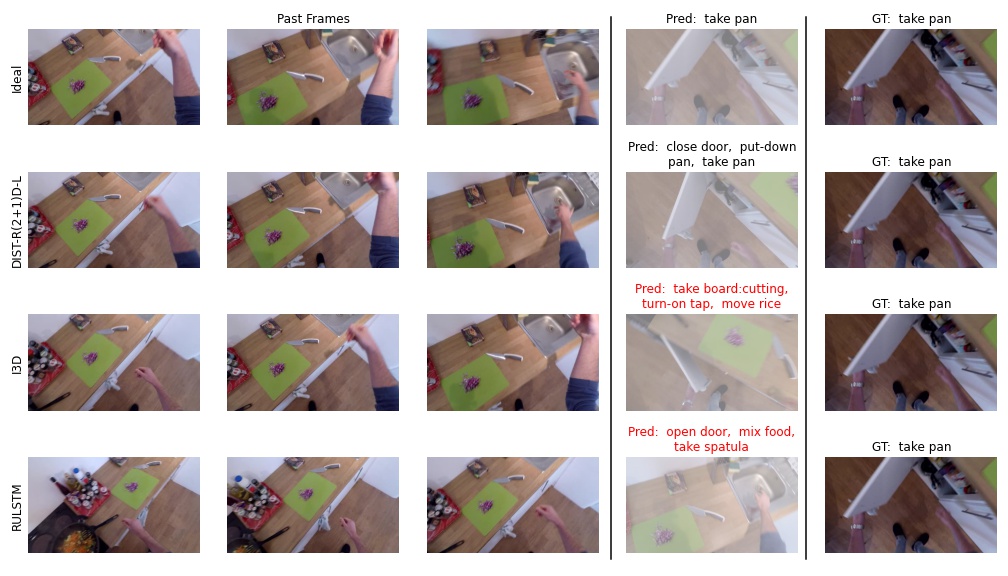}
    \caption{Two qualitative examples. In the streaming scenario, different methods base their prediction on different visual inputs due to their runtime. See text for discussion and the appendix for additional qualitative examples.}
    \label{fig:qualitative}
\end{figure*}

Figure~\ref{fig:qualitative} reports two qualitative examples comparing the performance of the proposed DIST-R(2+1)D-L, I3D and RULSTM on EPIC-KITCHENS-55. Due to their different runtimes, the methods base their predictions on different visual inputs. Each example is composed of four rows and five columns of images. The first three columns report three frames sampled from the video segment actually observed by the models. The fourth column reports the video frame appearing one second after the observed segment. This last frame is not observed by the model, but reported for reference only. The last column reports the first frame of the action to be anticipated. Also this frame is never observed by the model and reported only for reference.
The first row reports the frames which would be ideally observed by a model with zero runtime. As can be noted, the frames reported in the fourth and fifth columns are the same in this case. The second, third and fourth rows report the frames observed by the three considered methods.
We report the Ground Truth action (GT) above the images in the last column, whereas we report the Top-3 predictions above the images in the fourth column. Wrong predictions are reported in red.

In the first example (Figure~\ref{fig:qualitative}-top), both DIST-R(2+1)D and I3D  correctly predict the future action ``open fridge'' due to the fact that the fridge is actually observed at the end of the input sequence. Note that the input observed by DIST-R(2+1)D is more aligned to the beginning of the action and hence the correct prediction appears at rank-1, as opposed to rank-3 in the case of I3D.
In the second example (Figure~\ref{fig:qualitative}-bottom), both DIST-R(2+1)D-L and I3D observe a similar input sequence close to the ideal one. However, only DIST-R(2+1)D correctly anticipates the future ``take pan'' action. RULSTM observes a significantly different input segment and hence predicts ``open door'' which is likely to precede the ``take pan'' action.
Please see the appendix for additional qualitative examples.

\section{Conclusion}
\label{sec:conclusion}
We presented a novel streaming egocentric action evaluation scheme which, differently from current protocols, takes into account a challenging and realistic scenario in which action anticipation predictions have to be delivered in real-time under constrained computational resources.
The proposed protocol can be applied independently from the targeted device, which allows to develop and test algorithms accounting for a specific hardware.
Together with the new evaluation scheme, we proposed a benchmark of different computational methods which shows how current evaluation schemes offer a limited perspective, especially when methods have to be deployed on real hardware.
Based on the assumption that model runtime can severely affect accuracy in the considered streaming scenario, we proposed a knowledge distillation approach which can be used to greatly improve the performance of feed-forward 3D CNNs when used as action anticipation methods. 
We hope that the proposed evaluation scheme and investigation can help bring the attention of the research community on the practical issue of streaming egocentric action anticipation.

\appendix

\section{Implementation Details}
\label{sec:details}
In this section, we report the implementation details of the proposed and compared methods. We trained all models on a server equipped with $4$ NVIDIA V100 GPUs. 

\subsection{Proposed Method}
\label{sec:details_propsoed}
Regarding the distillation loss of Eq. (2) of the main paper, we set $\lambda_d=20$ and $\lambda_c=1$ in all our experiments.
We follow the procedure described in the main paper and train our model on both labeled and unlabeled data.
Specifically, we sample video pairs at all possible timestamps $t$ within all training videos, which accounts to about $3.5M$, $8M$, and $2.3M$ training examples on EPIC-KITCHENS-55, EPIC-KITCHENS-100, and EGTEA Gaze+ respectively.
To maximize the amount of labeled data, we consider a training example to be labeled if at least half of the frames of the future observation are associated to an action segment.
When more labels are included in a future observation (action segments may overlap), we associate it with the most frequent one.
If a past and a future observation contain the same action, we consider the example as unlabeled as we may be sampling in the middle of a long action.
Since training sets obtained in these settings are very large and partly redundant, we found all models to converge in one epoch.
We trained all models using the Adam optimizer~\cite{kingma2014adam} with a base learning rate of $1e-4$ and batch sizes equal to $28$ for the large variant, $80$ for medium variant, and $128$ for the small variant.
The teacher models are fine-tuned from Kinetics pre-trained weights provided in the PyTorch library.
During training of both teacher and student models, we perform random horizontal flip and resize the input video so that the shortest side is equal to $128$, $64$, or $32$ pixels, depending on the target resolution (large, medium and small respectively). After resizing the clip, we perform a random crop of the target resolution.
All these spatial augmentations are performed coherently on both video clips included in the training example.
During test, we remove the random horizontal clip and replace the random crop with a center crop.
After training, we obtain our final model by averaging the weights of the $5$ best-performing checkpoints.

\subsection{Compared Methods} 
\label{sec:details_compared}
We train RULSTM using the official code and pre-computed features provided by the authors\footnote{\url{https://github.com/fpv-iplab/rulstm}}~\cite{furnari2020rolling}. For ED, we follow the implementation and training scheme of~\cite{furnari2020rolling} and use the provided RGB and optical flow features. The TSN model has been trained using the Verb-Noun Marginal Cross Entropy loss proposed in~\cite{furnari2018leveraging} and adopting the suggested hyper-parameters. The LSTM baseline is trained using the same codebase of~\cite{furnari2020rolling} and same hyperparameters. The X3D-XS and SlowFast models are trained using the official code provided by the authors~\cite{feichtenhofer2019slowfast} and adopting the suggested parameters, finetuning from Kinetics-pretrained weights on 3 NVIDIA V100 GPUs. For I3D, we use a publicly available PyTorch implementation\footnote{https://github.com/piergiaj/pytorch-i3d}. The model is trained for $60$ epochs, with stochastic gradient descent and a base learning rate of $0.1$, which is multiplied by $0.1$ every $20$ epochs. The model is then finetuned from Kinetics-pretrained weights using 3 NVIDIA V100 GPUs and a batch size of $24$. We train the R(2+1)D-based models following the same parameters as the proposed approach. 

\subsection{Observation Times}
We set the methods' observation times using the parameters suggested by the authors in the related papers. Table~\ref{tab:obs_times} reports the adopted observation times.

 \begin{table}[t]
 \caption{Observation times of the compared methods in seconds.}
 \label{tab:obs_times}
 \centering
 \adjustbox{max width=\linewidth}{
 \setlength{\tabcolsep}{2.5pt}
\begin{tabular}{lr}
\hline
\textbf{METHOD}                            & \textbf{$\bf \tau_o$} \\ \hline
RULSTM \citep{furnari2020rolling}          & 2.75                               \\
ED \citep{gao2017red}                      & 4                                  \\ \hline
I3D \citep{carreira2017quo}                & 2.14                               \\
SlowFast \citep{feichtenhofer2019slowfast} & 2.14                               \\
X3D-XS \citep{feichtenhofer2020x3d}        & 1.07                               \\
R(2+1)D-S \citep{tran2018closer}           & 1.07                               \\
R(2+1)D-M \citep{tran2018closer}           & 1.07                               \\
R(2+1)D-L \citep{tran2018closer}           & 1.07                               \\ \hline
LSTM \citep{gers1999learning}              & 2.75                               \\
TSN \citep{li2020towards}                  & 1                                  \\ \hline
DIST-R(2+1)D-S                                              & 1.07                               \\
DIST-R(2+1)D-M                                              & 1.07                               \\
DIST-R(2+1)D-L                                              & 1.07                               \\ \hline
LSTM+DIST-R(2+1)D-S                                         & 2.75                               \\
LSTM+DIST-R(2+1)D-M                                         & 2.75                               \\
LSTM+DIST-R(2+1)D-L                                         & 2.75                               \\ \hline
\end{tabular}}
\end{table}

\section{Qualitative Examples}
\label{sec:qualitative}
\begin{figure*}
    \centering
    \includegraphics[width=\linewidth]{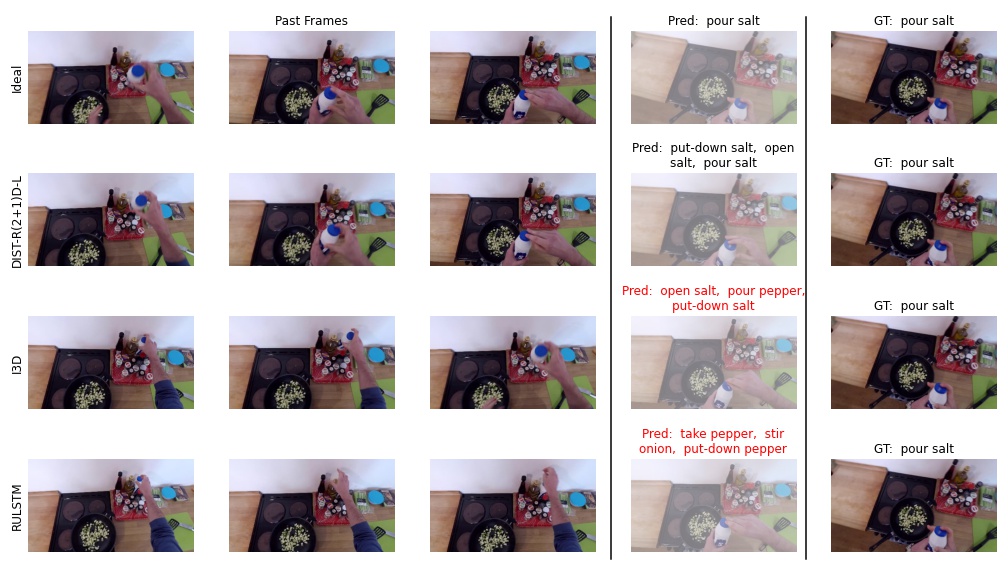}
    \caption{Qualitative examples.}
    \label{fig:qualitative_1}
\end{figure*}

\begin{figure*}
    \centering
    \includegraphics[width=\linewidth]{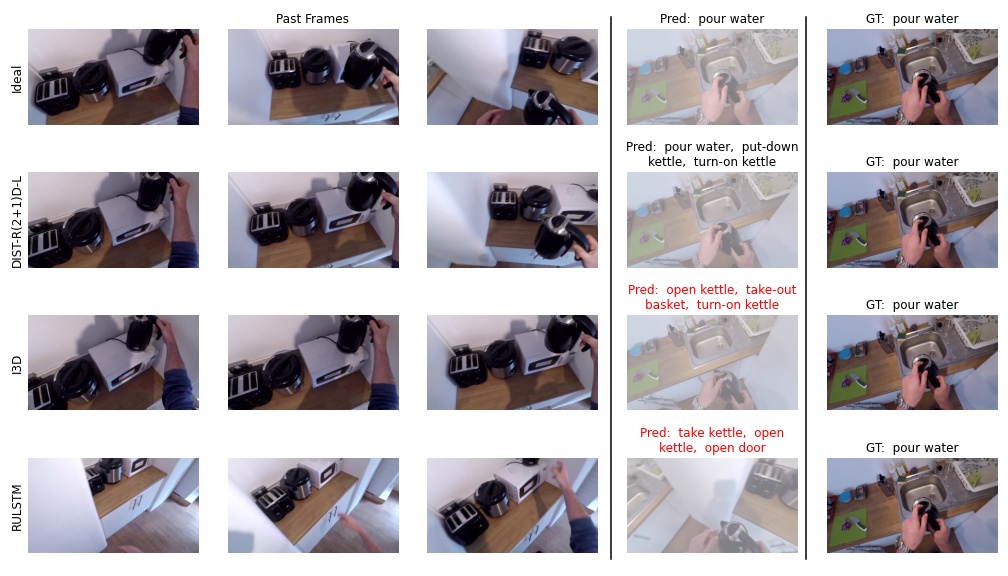}
    \caption{Qualitative examples.}
    \label{fig:qualitative_2}
\end{figure*}

\begin{figure*}
    \centering
    \includegraphics[width=\linewidth]{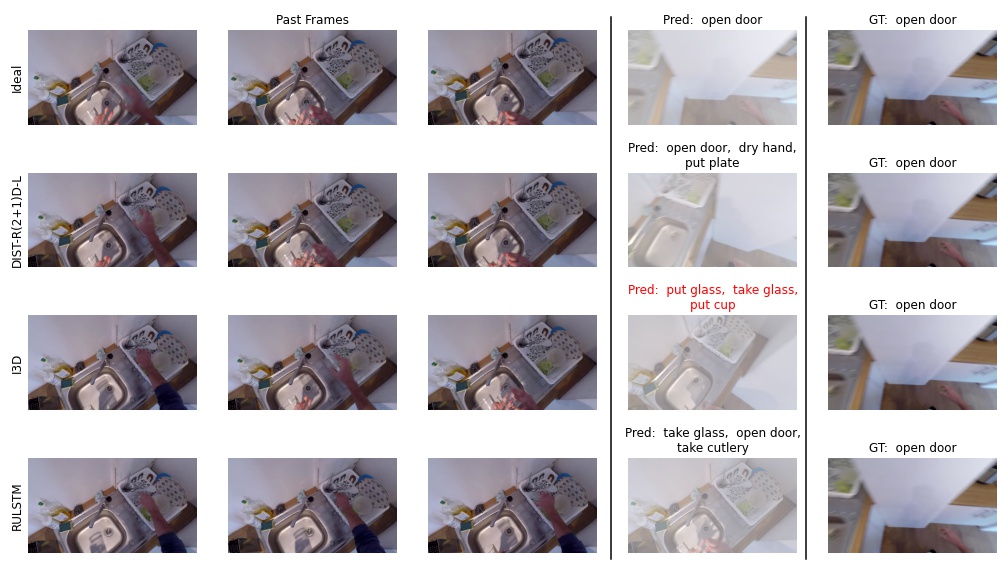}
    \caption{Qualitative examples.}
    \label{fig:qualitative_3}
\end{figure*}

\begin{figure*}
    \centering
    \includegraphics[width=\linewidth]{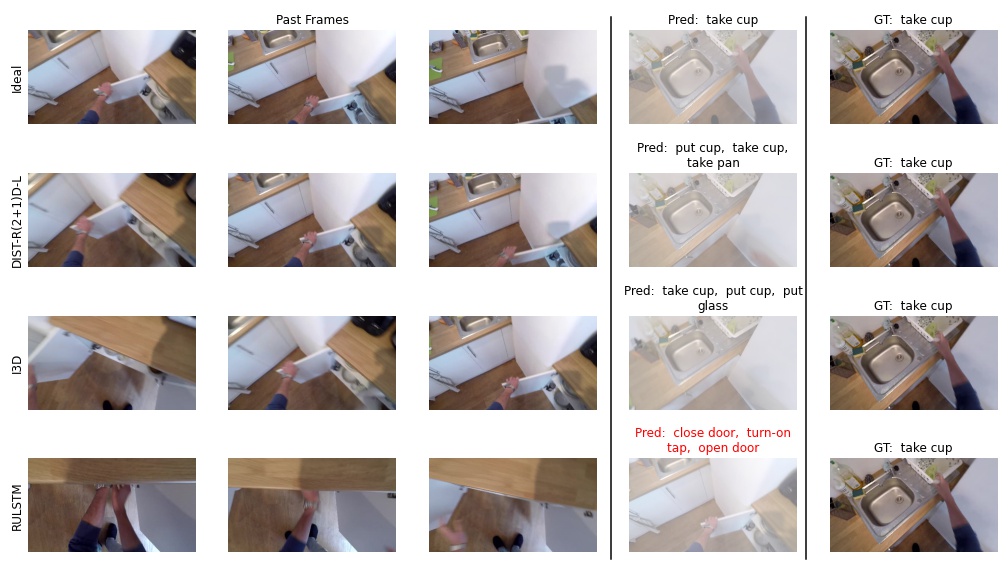}
    \caption{Qualitative examples.}
    \label{fig:qualitative_4}
\end{figure*}

\begin{figure*}
    \centering
    \includegraphics[width=\linewidth]{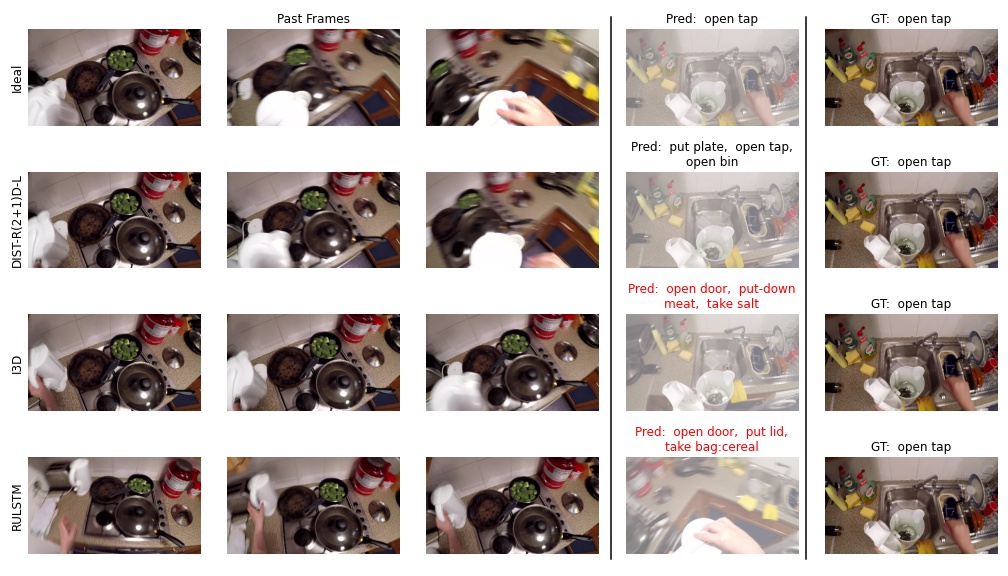}
    \caption{Qualitative examples.}
    \label{fig:qualitative_5}
\end{figure*}

\begin{figure*}
    \centering
    \includegraphics[width=\linewidth]{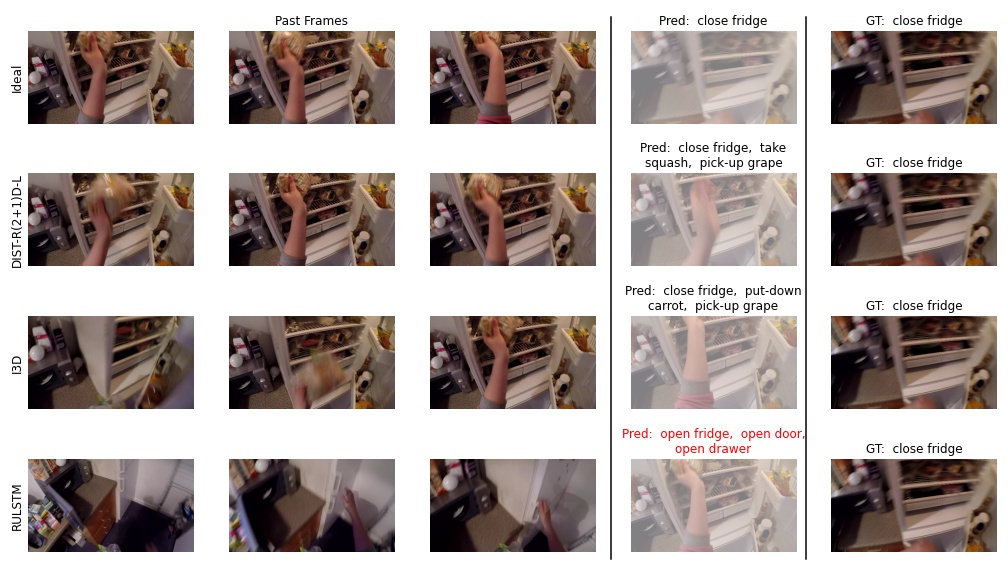}
    \caption{Qualitative examples.}
    \label{fig:qualitative_6}
\end{figure*}

\begin{figure*}
    \centering
    \includegraphics[width=\linewidth]{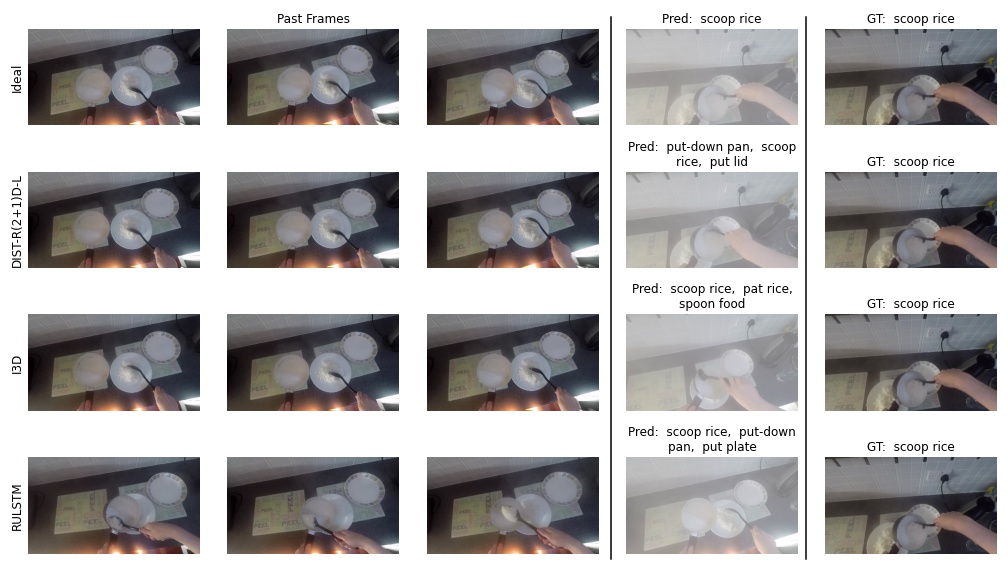}
    \caption{Qualitative examples.}
    \label{fig:qualitative_7}
\end{figure*}

\begin{figure*}
    \centering
    \includegraphics[width=\linewidth]{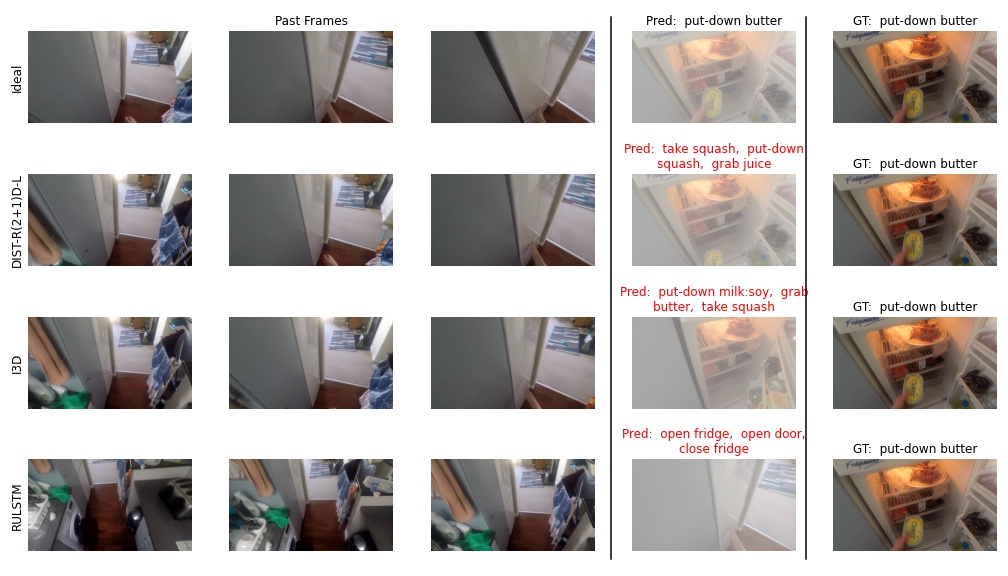}
    \caption{Qualitative examples.}
    \label{fig:qualitative_8}
\end{figure*}

\begin{figure*}
    \centering
    \includegraphics[width=\linewidth]{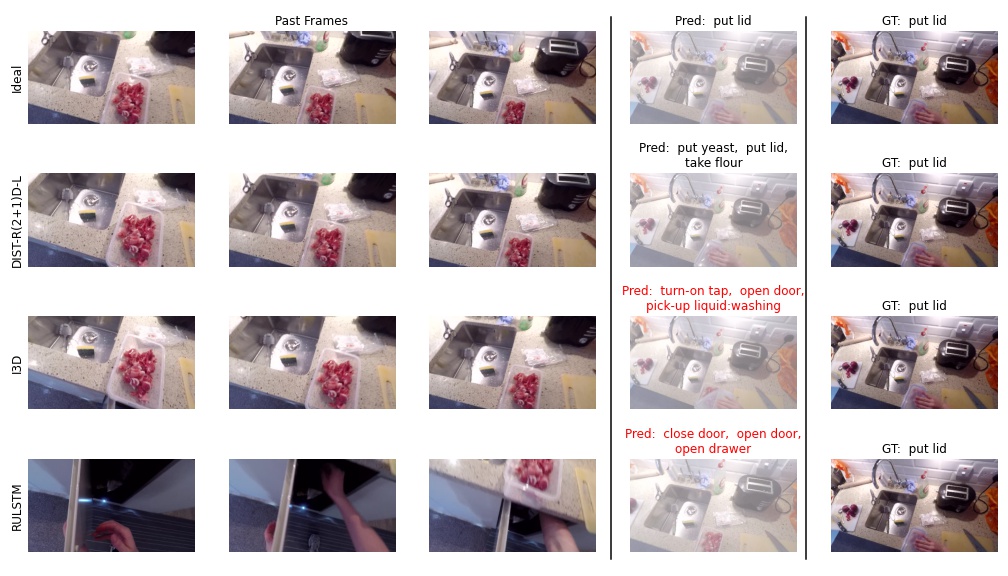}
    \caption{Qualitative examples.}
    \label{fig:qualitative_9}
\end{figure*}

\begin{figure*}
    \centering
    \includegraphics[width=\linewidth]{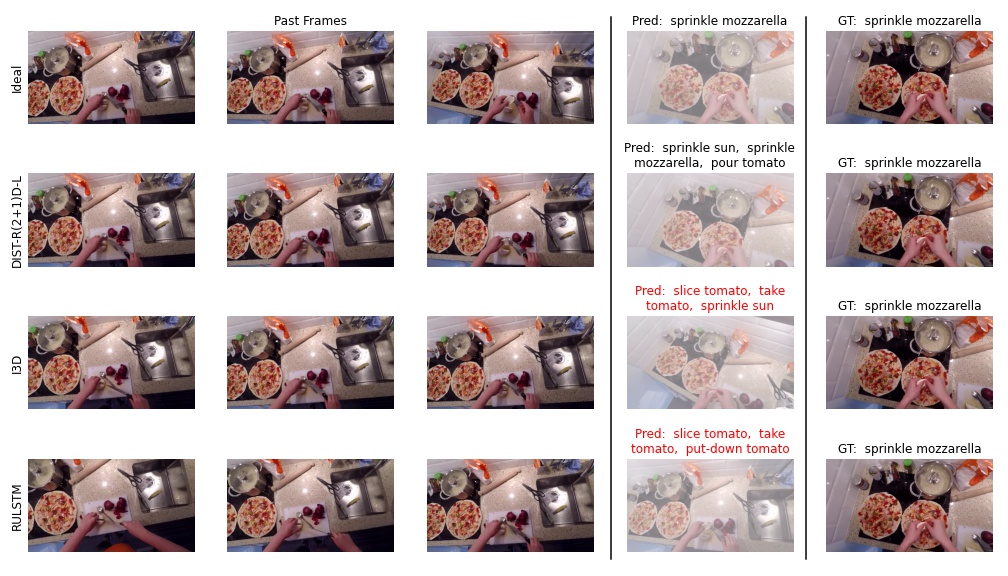}
    \caption{Qualitative examples.}
    \label{fig:qualitative_10}
\end{figure*}

\begin{figure*}
    \centering
    \includegraphics[width=\linewidth]{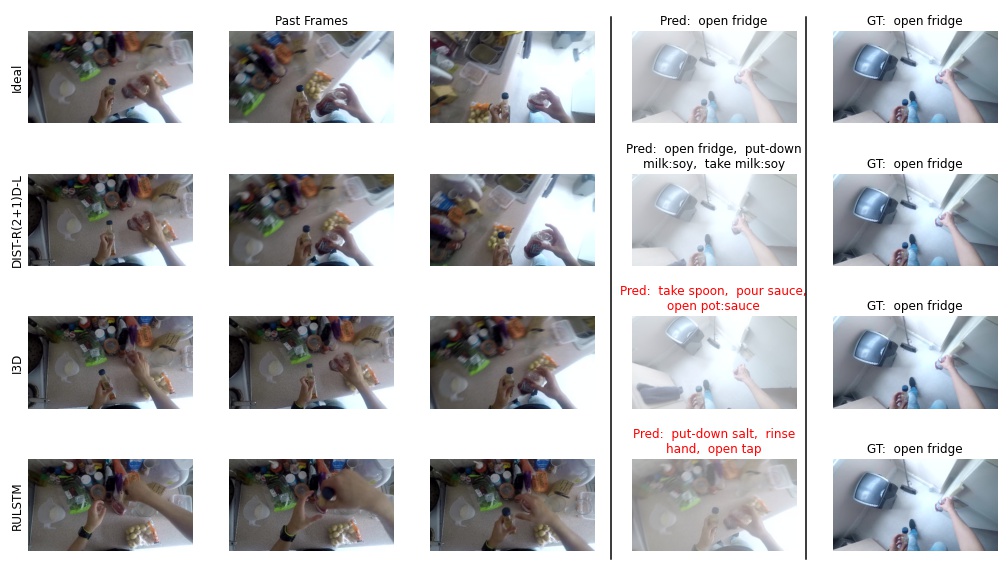}
    \caption{Qualitative examples.}
    \label{fig:qualitative_11}
\end{figure*}

\begin{figure*}
    \centering
    \includegraphics[width=\linewidth]{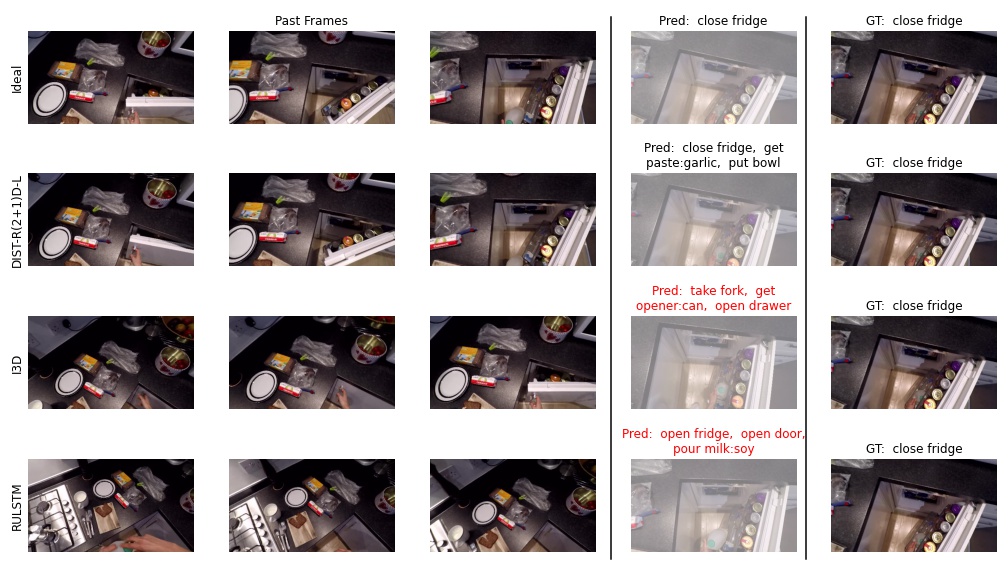}
    \caption{Qualitative examples.}
    \label{fig:qualitative_12}
\end{figure*}

\begin{figure*}
    \centering
    \includegraphics[width=\linewidth]{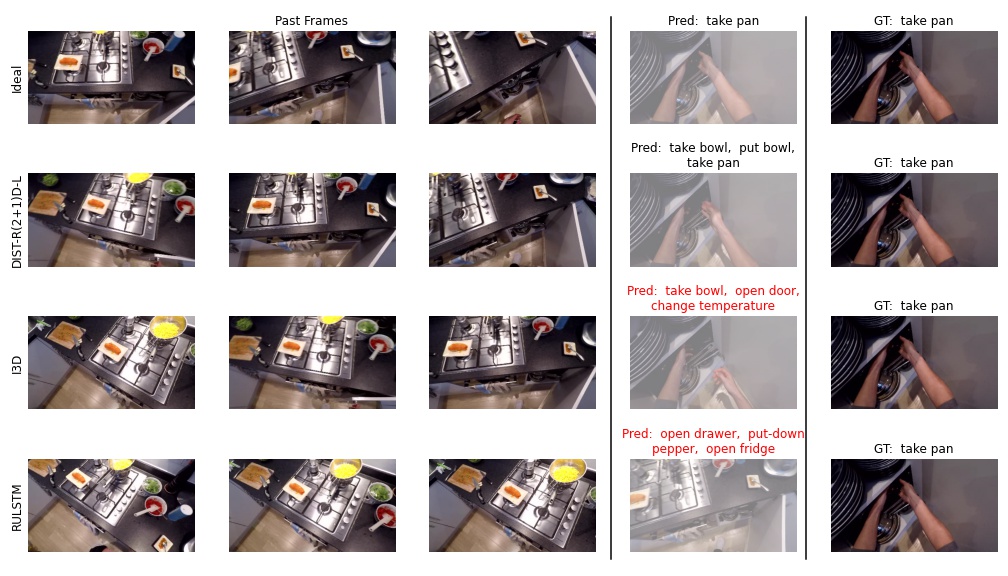}
    \caption{Qualitative examples.}
    \label{fig:qualitative_13}
\end{figure*}

\begin{figure*}
    \centering
    \includegraphics[width=\linewidth]{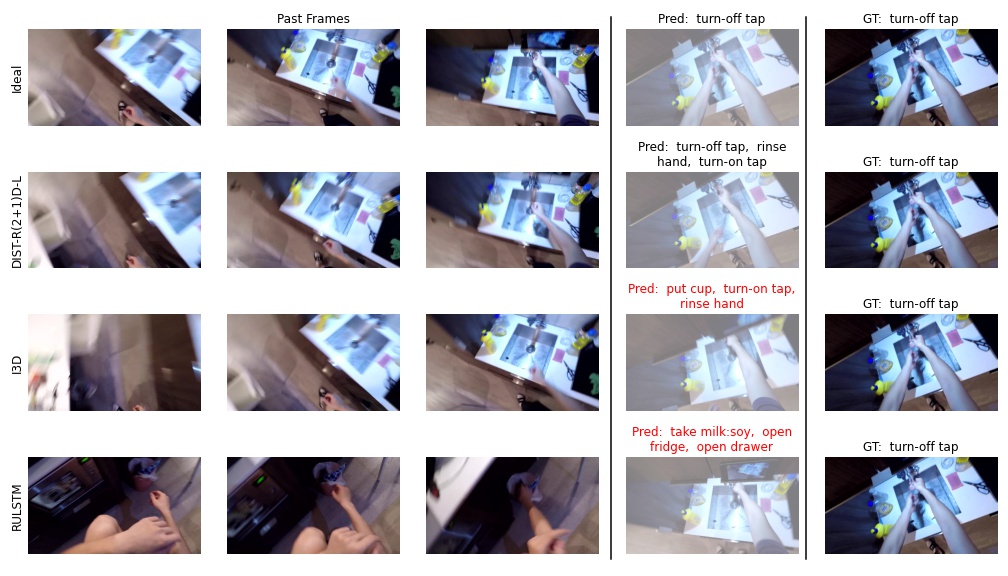}
    \caption{Qualitative examples.}
    \label{fig:qualitative_14}
\end{figure*}

\begin{figure*}
    \centering
    \includegraphics[width=\linewidth]{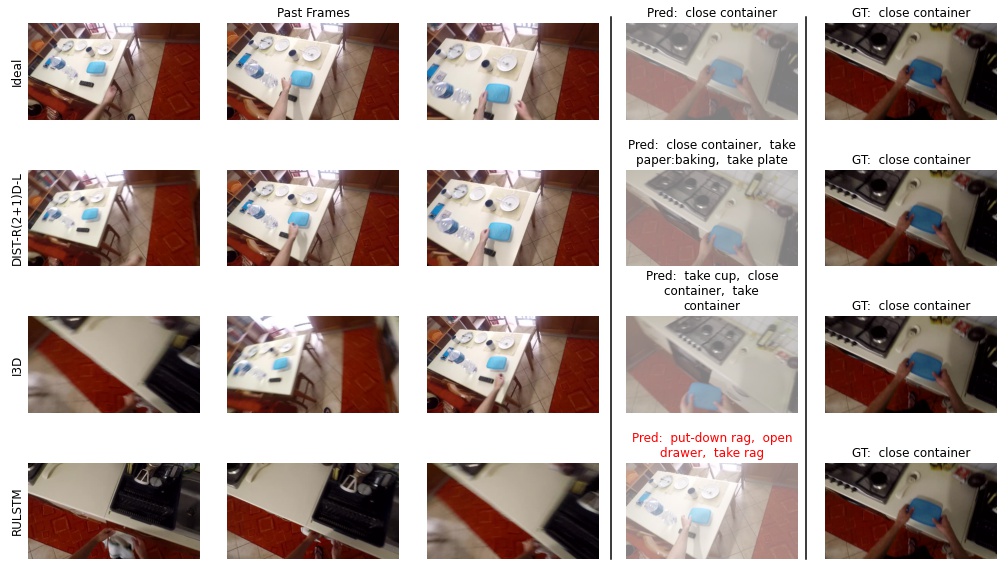}
    \caption{Qualitative examples.}
    \label{fig:qualitative_15}
\end{figure*}

\begin{figure*}
    \centering
    \includegraphics[width=\linewidth]{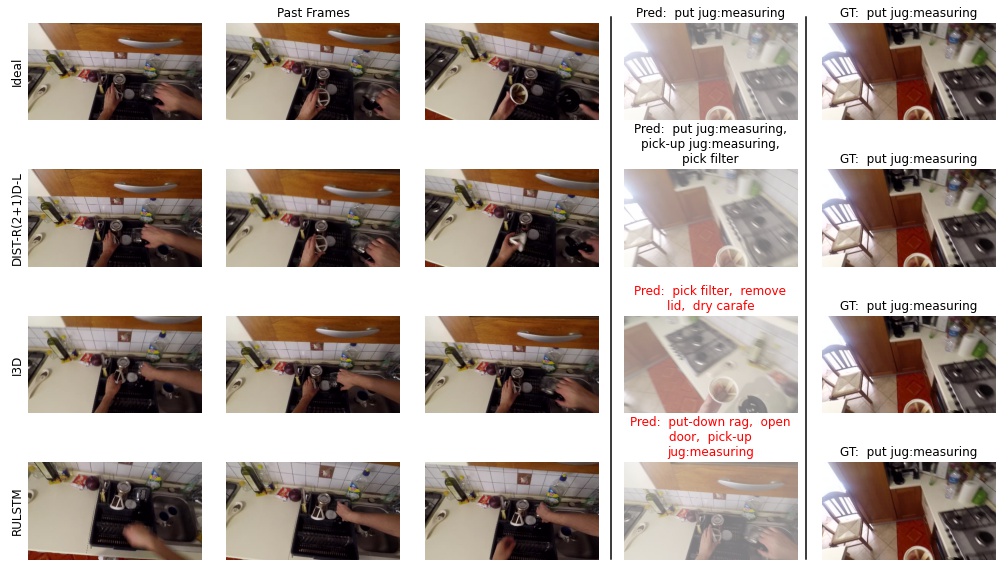}
    \caption{Qualitative examples.}
    \label{fig:qualitative_16}
\end{figure*}

\begin{figure*}
    \centering
    \includegraphics[width=\linewidth]{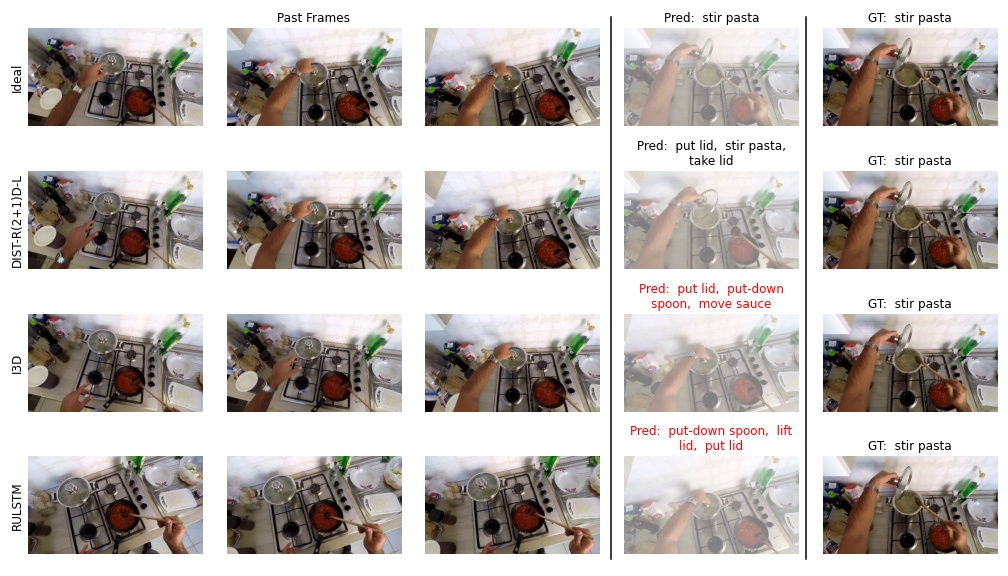}
    \caption{Qualitative examples.}
    \label{fig:qualitative_17}
\end{figure*}

\begin{figure*}
    \centering
    \includegraphics[width=\linewidth]{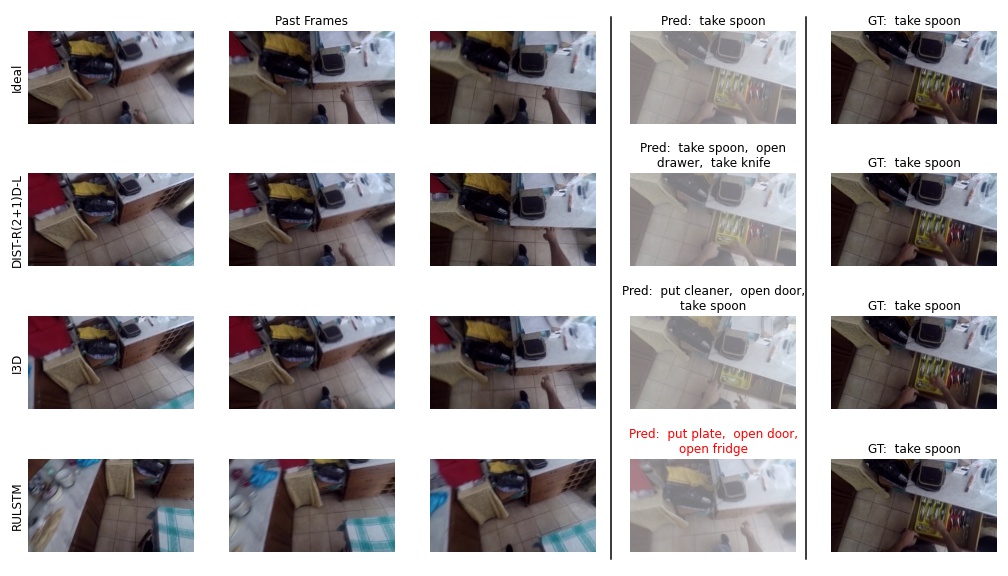}
    \caption{Qualitative examples.}
    \label{fig:qualitative_18}
\end{figure*}

Figures~\ref{fig:qualitative_1}-\ref{fig:qualitative_18} report additional qualitative examples. 
As can be noted, RULSTM usually observes a significantly different input sequence, which makes predictions less effective (e.g., Figures~\ref{fig:qualitative_4},~\ref{fig:qualitative_6},~\ref{fig:qualitative_8},~\ref{fig:qualitative_9},~\ref{fig:qualitative_14},~\ref{fig:qualitative_15},~\ref{fig:qualitative_18}).
Despite DIST-R(2+1)D-L and I3D tend to observe similar inputs, the former tends to make better predictions than the latter (e.g., Figures~\ref{fig:qualitative_1},~\ref{fig:qualitative_2},~\ref{fig:qualitative_3},~\ref{fig:qualitative_5},~\ref{fig:qualitative_9},~\ref{fig:qualitative_10},~\ref{fig:qualitative_16},~\ref{fig:qualitative_17}).
Methods with different runtimes tend to observe the same input video and make similar predictions in the case of stationary scenes (e.g., Figure~\ref{fig:qualitative_7}).


\section*{Acknowledgment}
This research has been supported by Future Artificial Intelligence Research (FAIR) – PNRR MUR Cod. PE0000013 - CUP: E63C22001940006, and by MIUR AIM - Attrazione e Mobilit\`a Internazionale Linea 1 - AIM1893589 - CUP: E64118002540007.

\bibliographystyle{model2-names}
\bibliography{refs}

\end{document}